\documentclass[sigconf,nonacm]{acmart}
\usepackage{multirow}
\usepackage{subcaption}
\usepackage{graphicx}
\usepackage{balance}
\usepackage{mathtools}
\usepackage{url}
\usepackage{amsfonts}
\usepackage{textcomp}
\usepackage{color}
\usepackage{enumitem}
\usepackage{tikz}
\usepackage{pgfplots,pgfplotstable}
\usepackage{pdfrender}
\usepackage{amsmath}
\usepackage{bbm}
\usepackage{xcolor} 
\usepackage[table]{xcolor}
\usepackage{booktabs}
\usepackage{algorithm}
\usepackage{algpseudocode}
\usepackage{bm} 
\usepackage{caption}
\usepackage{adjustbox}
\definecolor{selfadvgreen}{HTML}{1B7F3A}
\definecolor{ourpink}{HTML}{D81B60}

\definecolor{gainblue}{HTML}{0800EA}
\definecolor{gainred}{HTML}{D10300}
\definecolor{gaingrey}{HTML}{9B9D9D}

\definecolor{uniform}{HTML}{11DD11}
\definecolor{uniform-sans}{HTML}{FFA34D}
\definecolor{uniform-sans-rw}{HTML}{917600}
\definecolor{selfadv}{HTML}{086908}
\definecolor{selfadv-sans}{HTML}{D76DAA}
\definecolor{selfadv-sans-rw}{HTML}{9800CA}
\definecolor{erdns}{HTML}{1F77B4}
\definecolor{eans}{HTML}{1100FF}
\definecolor{condns}{HTML}{000000}
\definecolor{ours}{HTML}{FF0000}

\newcommand{\VN}[1]{\textcolor{venue}{#1}}

\newcommand{\method}{\textsf{RADAR}}
\newcommand{\codeurl}{\url{https://anonymous.4open.science/r/KGENS-1EB6/}}

\newcommand{\legstroke}[2]{%
    \tikz[baseline=-0.5ex]{%
        \draw[
            draw=#1,
            line width=1.5pt,
            line cap=butt,
            #2
        ] (0,0) -- (0.85,0);
    }%
    \hspace{2pt}%
}

\newcommand{\methodlegend}{%
    \begingroup
    \setlength{\tabcolsep}{3pt}
    \renewcommand{\arraystretch}{1.15}
    \noindent\resizebox{\columnwidth}{!}{%
        \begin{tabular}{@{}lllll@{}}
            \legstroke{uniform}{dotted} Uniform &
            \legstroke{uniform-sans}{dash pattern=on 2pt off 1.5pt} Uniform SANS &
            \legstroke{uniform-sans-rw}{dash pattern=on 8pt off 2.5pt} Uniform RW-SANS &
            \legstroke{erdns}{dashed} ERDNS &
            \legstroke{condns}{dash pattern=on 1pt off 3pt} ConDNS \\

            \legstroke{selfadv}{dash dot} Self-Adv &
            \legstroke{selfadv-sans}{dash pattern=on 4pt off 1.5pt on 1pt off 1.5pt on 1pt off 1.5pt} Self-Adv SANS &
            \legstroke{selfadv-sans-rw}{dash pattern=on 7pt off 2pt on 2pt off 2pt} Self-Adv RW-SANS &
            \legstroke{eans}{dash pattern=on 4pt off 1.5pt on 4pt off 1.5pt on 1pt off 1.5pt} EANS &
            \legstroke{ours}{solid} \textbf{\method} \\
        \end{tabular}%
    }%
    \par\vspace{0.1cm}
    \endgroup
}

\definecolor{rotatecolor}{HTML}{E84A4A}
\definecolor{transecolor}{HTML}{FF9F40}
\definecolor{complexcolor}{HTML}{2CB34A}
\definecolor{distmultcolor}{HTML}{1F77B4}

\newcommand{\modellegend}{%
  \makebox[\linewidth][c]{%
  \footnotesize
    \tikz[baseline=-0.5ex]{
      \draw[rotatecolor, line width=0.7pt] (0,0)--(0.6,0);
      \fill[rotatecolor] (0.3,0) circle (2.0pt);
    }~RotatE%
    \hspace{1.2em}%
    \tikz[baseline=-0.5ex]{
      \draw[transecolor, line width=0.7pt] (0,0)--(0.6,0);
      \draw[transecolor, line width=0.7pt]
        (0.23,-0.07)--(0.37,0.07)
        (0.23,0.07)--(0.37,-0.07);
    }~TransE%
    \hspace{1.2em}%
    \tikz[baseline=-0.5ex]{
      \draw[complexcolor, line width=0.7pt] (0,0)--(0.6,0);
      \fill[complexcolor]
        (0.23,-0.07) rectangle (0.37,0.07);
    }~ComplEx%
    \hspace{1.2em}%
    \tikz[baseline=-0.5ex]{
      \draw[distmultcolor, line width=0.7pt] (0,0)--(0.6,0);
      \fill[distmultcolor]
        (0.3,0.09)--(0.38,0)--(0.3,-0.09)
        --(0.22,0)--cycle;
    }~DistMult%
  }%
}

\newcommand{\E}{\mathcal{E}}

\newcommand{\R}{\mathcal{R}}
\newcommand{\T}{\mathcal{T}}

\newcommand{\x}{\textbf{x}}
\newcommand{\sa}{Self-Adv}

\definecolor{myblue}{HTML}{6CB2F8}

\definecolor{myHexBlue}{HTML}{1F3782}
\definecolor{myHexRed}{HTML}{8C0009}

\definecolor{first}{HTML}{BDDDFC}
\definecolor{second}{HTML}{F2F2F2}

\definecolor{venue}{HTML}{385B86}

\definecolor{legend2}{HTML}{577590}
\definecolor{legend1}{HTML}{DA9474}
\newcommand{\legendbar}[1]{\textcolor{#1}{\rule[-0.15ex]{0.9em}{1.4ex}}}

\AtBeginDocument{
    \hypersetup{
        colorlinks=true,    
        linkcolor=myHexRed,     
        citecolor=myHexBlue,    
        filecolor=myHexBlue,  
        urlcolor=myHexBlue       
    }
}


\title[How Many and How Much? Contrastive Learning with Negative Samples for Knowledge Graph Embedding]{How Many and How Much? Contrastive Learning with \\Negative Samples for Knowledge Graph Embedding}

\author{Sooho Moon}
\affiliation{%
  \institution{Chung-Ang University}
  \city{Seoul}
  \country{Korea}}
\email{moonwalk725@cau.ac.kr}

\author{Yunyong Ko}
\authornote{Corresponding author}
\affiliation{%
  \institution{Chung-Ang University}
  \city{Seoul}
  \country{Korea}}
\email{yyko@cau.ac.kr}
\orcid{0000-0003-1283-4697}

\begin{document}

\begin{abstract}
Knowledge graph embedding (KGE), a fundamental technique for knowledge-centric Web and KG-augmented AI applications, 
is commonly trained through contrastive learning with negative samples to learn discriminative representations.
Despite its empirical success, two crucial questions regarding negative samples remain underexplored:
\textbf{(Q1)} \textit{how many negative samples should be used for each positive sample} and \textbf{(Q2)} \textit{how much should each negative sample contribute to learning?}
To answer these questions, we propose \textbf{{\method}}, a novel contrastive learning framework for KGE models with two effective strategies, respectively:
(1) \textit{Risk-aware negative budget allocation} quantifies false-negative sampling risk by modeling an open-world KG as a stochastic partial observation of its closed-world counterpart and adaptively assigns a negative budget to each positive sample based on its risk;
(2) \textit{Distribution-calibrated negative weighting} 
assigns negative weights by calibrating the weighting strength for each positive based on two complementary characteristics of the negative-score distribution: dispersion and concentration.
Extensive experiments on 3 real-world KGs with 4 KGE models reveal that {\method} achieves the best KGE performance on average across all model--KG combinations and reduces sampled false negatives by over $90\%$ compared with conventional fixed-budget methods.
Our code and datasets are available at {\codeurl}.
\end{abstract}

\maketitle

\section{Introduction}\label{sec:intro}
The Web serves as an abundant source of information about real-world entities and their complex relationships.
A knowledge graph (KG) organizes such information into structured relational knowledge, where each fact is represented as a triple, \textit{(head entity, relation, tail entity)}, such as Freebase~\cite{bollacker2008freebase} and YAGO~\cite{suchanek2007yago}.
By making relational knowledge explicit, KGs have become an important foundation for a wide range of knowledge-centric applications, including Web search~\cite{noy2019kg-web,dong2014kg-web}, recommender systems~\cite{wang2019kge-rec,wang2019kg-rec}, question answering~\cite{huang2019kgqa-www,yasunaga2021kgqa-acl}, and even drug discovery~\cite{wang2021kg-drug,zeng2022drug}.

Beyond explicit relational knowledge, KG embedding (KGE)~\cite{bordes2013translating,wang2014transh,trouillon2016complex,yang2015embeddingentitiesrelationslearning,sun2019rotate,xiao2024ivr,li2025decoupled} aims to learn representations of entities and relations to capture latent relational patterns in a KG.
The learned representations can then be utilized for diverse downstream tasks.
Most notably, KG completion (KGC)~\cite{bordes2013translating,yang2015embeddingentitiesrelationslearning,trouillon2016complex,zhu2021neural,zhang2022knowledge} aims to discover missing facts in a KG, 
where KGE estimates their plausibility based on the learned representations of the involved entities and relations.
In addition to KGC, learned representations can serve as structural features for downstream tasks such as recommendation and enzyme prediction~\cite{song2026kg-chem, wang2019kge-rec}.
They can also support emerging KG-augmented AI applications, such as LLM reasoning with graph-based RAG, by enabling the retrieval of relevant knowledge based on its similarity to a query in the latent representation space~\cite{luo2026gfm,yasunaga2022lang-kg}.

\begin{figure}[t]
\centering
\includegraphics[width=0.99\linewidth]{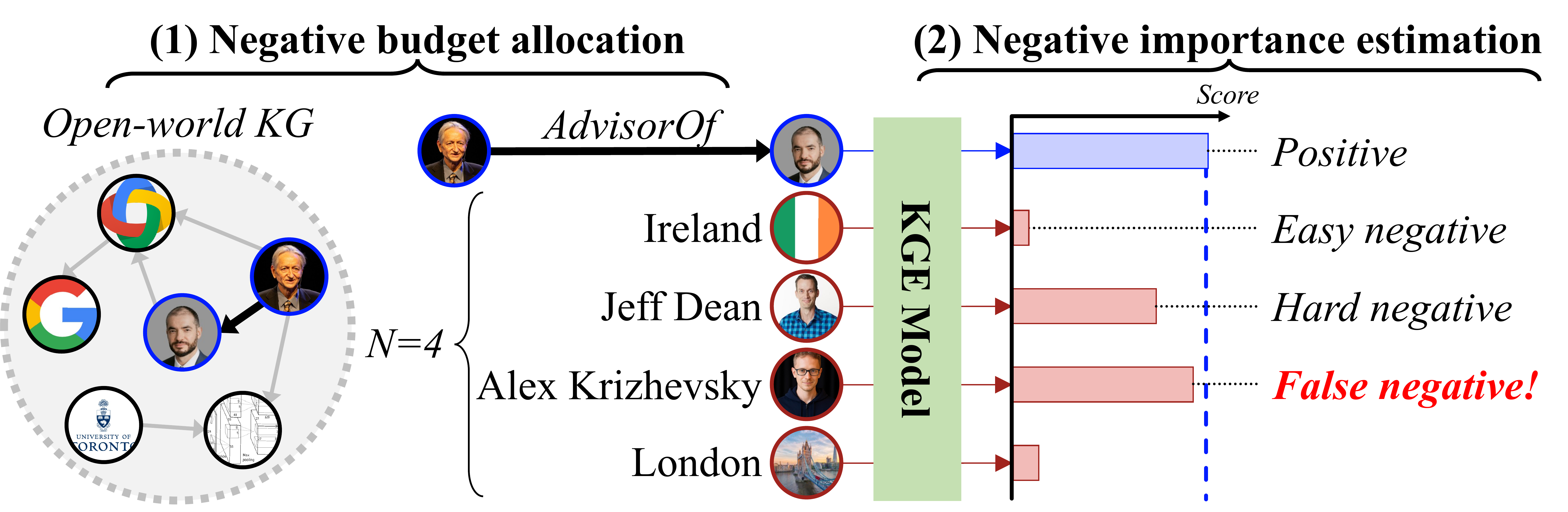}
\caption{Process of contrastive learning with negative samples for knowledge graph embedding.}
\vspace{-5mm}
\label{fig:kge-process}
\end{figure}

KGE models are generally trained through \textit{contrastive learning with negative samples}~\cite{yang2024ns-survey,gutmann2010nce,zhang2019nscaching,sun2019rotate}, which learns to assign higher scores to positive samples (i.e., true facts) and lower scores to negative samples (i.e., false facts), as illustrated in Figure~\ref{fig:kge-process}.
Since false facts are not explicitly available in a KG, negative samples are typically constructed by corrupting a positive sample $(h,r,t)$, i.e., replacing its head or tail entity with another entity such that the resulting triple is not observed in the KG.
This learning paradigm enables KGE models to learn discriminative representations that distinguish true facts from false ones.

\begin{figure}[t]
    \centering
    \includegraphics[width=0.46\linewidth]{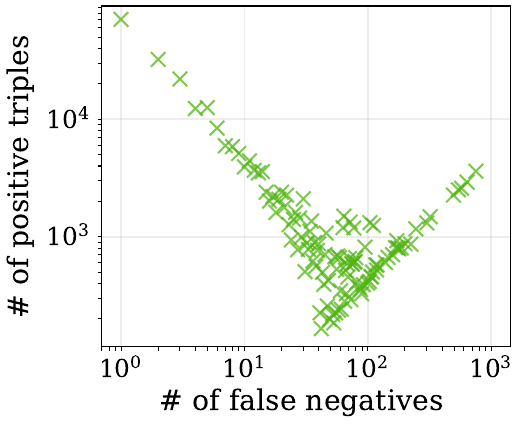}
    \includegraphics[width=0.46\linewidth]{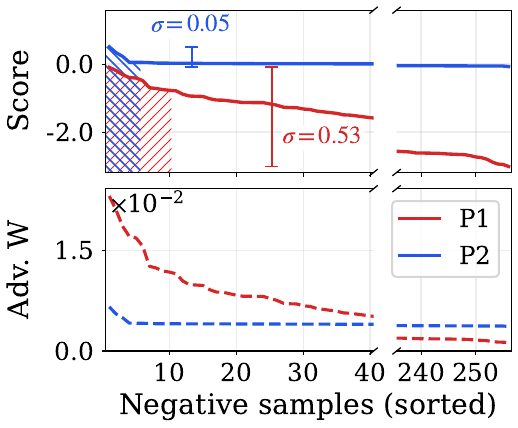}
    \vspace{-1mm}
    \caption{Observations: (a) Varying numbers of false negatives across positive triples on FB15k237~\cite{toutanova2015observed}, and (b) negative scores and their weights computed by \sa~\cite{sun2019rotate} for two positive triples (P1 and P2) with different score dispersion ($\sigma$) and concentration (shaded: top-10\% negative scores).}\label{fig:observation}
    \vspace{-5mm}
\end{figure}

\vspace{0.5mm}
\noindent
\textbf{Challenges.}
Despite its empirical success, 
the following two challenges regarding negative samples still remain underexplored:

\textbf{(C1) Negative budget allocation.}
For each positive triple, KGE models are typically trained with a \textit{fixed} number of negative samples (e.g., 128 or 256)~\cite{zhang2019nscaching,sun2019rotate,wang2014transh}.
However, determining a proper negative budget is non-trivial since real-world KGs follow the open-world assumption (OWA)~\cite{yang2022rethinking,galarraga2017kgcompleteness,moon2026generalized}.
Under the OWA, an unobserved fact is not necessarily false, implying that some negative samples drawn from unobserved facts may actually be true---i.e., \textit{false negatives} (FN).
This introduces an inherent trade-off: sampling more negatives increases the risk of FNs, whereas sampling too few negatives may limit the supervision available for contrastive learning.
More importantly, we observe that the number of FNs varies significantly across positive triples (see Figure~\ref{fig:observation}(a)), 
where FNs for each training positive $(h,r,t)$ are the triples in the validation and test sets that share the same $(h,r)$ or $(r,t)$ pair (i.e., anchor) but are absent from the training set.
This raises a fundamental question: \textit{how many negatives should be used for each positive according to its false-negative risk?}
Answering this question is inherently challenging, as accurately estimating such risk requires closed-world knowledge of whether unobserved facts are actually true, which is unavailable under the OWA.

\textbf{(C2) Negative importance estimation}.
Once negatives are sampled, another challenge is to determine \textit{how much each negative should contribute to learning}.
Rather than treating all negatives equally,
self-adversarial weighting (\sa)~\cite{sun2019rotate} estimates their relative hardness based on model scores: higher-scoring negatives are considered harder to distinguish from a positive and are thus assigned larger weights, allowing them to contribute more strongly to learning.
Our analytical decomposition of {\sa}, however, reveals that its weighting strength is determined solely by the \textit{dispersion} of negative scores:
more dispersed scores induce larger weight differences, whereas less dispersed scores yield more uniform weights.
Such dispersion-based scaling can overlook another critical characteristic of the score distribution---\textit{how strongly high scores are concentrated on a few negatives}.
As illustrated in Figure~\ref{fig:observation}(b), where P1 and P2 denote two positives,
the top-10\% negative scores of P1's negatives are less concentrated (red shaded), yet their larger dispersion ($\sigma=0.53$) results in larger weight differences.
In contrast, the top-10\% negative of P2's negatives are concentrated on fewer negatives (blue shaded), but their smaller dispersion ($\sigma=0.05$) leads to lower weights.
Such an oversight of score concentration raises another fundamental question: 
\textit{how much should each negative contribute to learning considering both the dispersion and concentration of negative scores?}

\vspace{1mm}
\noindent
\textbf{Our work.}
To address these challenges, in this paper,
we propose \textbf{{\method}}, a novel contrastive learning framework for KGE that integrates (1) \textbf{\underline{R}}isk-\textbf{\underline{A}}ware negative budget allocation and (2) \textbf{\underline{D}}istribution-c\textbf{\underline{A}}lib\textbf{\underline{R}}ated negative weighting.

\textbf{(1) Risk-aware negative budget allocation}.
We first analyze the gap between an open-world KG and its ideal closed-world counterpart, where all true facts are observed.
Specifically, we interpret an open-world KG as a stochastic partial observation of its corresponding closed-world KG.
Based on this interpretation, we quantify the risk of sampling false negatives for each positive triple as a probability and derive its relationship to the observed degree.
We then adaptively determine the negative budget for each positive triple, allocating fewer negatives to higher-risk triples and more negatives to lower-risk ones.

\textbf{(2) Distribution-calibrated negative weighting}. 
To reflect score concentration in negative weighting, we then design an entropy-based measure that quantifies the score concentration of negatives.
Specifically, the measure is computed from the entropy of a probability distribution derived from standardized negative scores, 
with larger values indicating that high scores are concentrated among fewer negatives.
We observe that score dispersion and concentration are not strongly correlated (see Figure~\ref{fig:disp-concentration}), 
implying that score concentration provides complementary information that is not captured solely by dispersion.
Based on this observation, we jointly consider the dispersion and concentration of negative scores to determine the weighting strength for each positive triple, thereby calibrating its negative weights according to both characteristics of the negative score distribution.


\vspace{0.5mm}
\noindent
\textbf{Contributions.}
The main contributions of our work are as follows:
\begin{itemize}[leftmargin=10pt]
\item \textbf{Challenges.} We identify two underexplored challenges in contrastive learning with negative samples for KGE: \textbf{(C1)} negative budget allocation and \textbf{(C2)} negative importance estimation.
\item \textbf{Framework.} We propose \textbf{{\method}}, a novel contrastive learning framework that addresses the two challenges through (1) \textit{risk-aware negative budget allocation} and (2) \textit{distribution-calibrated negative weighting}, respectively.
\item \textbf{Evaluation.} Via extensive experiments on 3 real-world KGs with 4 KGE models, we demonstrate that {\method} (1) achieves the best KGE performance on average across all model--dataset combinations and (2) reduces sampled false negatives by over $90\%$ compared with existing fixed-budget methods.
\end{itemize}

\section{Related Work}\label{sec:related}

\noindent
\textbf{Knowledge graph embedding}.
Knowledge graph embedding (KGE) models~\cite{wang2014transh,bordes2013translating,yang2015embeddingentitiesrelationslearning,trouillon2016complex,sun2019rotate,balazevic2019tucker} aim to learn latent representations of entities and relations to capture relational patterns in a KG.
\textit{Translation-based models} interpret a relation as a transformation between entity embeddings. TransE~\cite{bordes2013translating} models each relation as a translation, $\textbf{h}+\textbf{r}\approx\textbf{t}$, 
while RotatE~\cite{sun2019rotate} models relations as rotations in complex space. 
HousE~\cite{li2022house} uses Householder transformation for rotation and projection to capture complex relations.
DGS~\cite{iyer2022dgs} models cyclic and hierarchical relations using spherical and hyperbolic embedding spaces, respectively.
\textit{Tensor decomposition models} represent a KG as a three-way tensor and factorize it using entity and relation representations. DistMult~\cite{yang2015embeddingentitiesrelationslearning} employs a diagonal bilinear scoring function, ComplEx~\cite{trouillon2016complex} extends it to the complex domain, and TuckER~\cite{balazevic2019tucker} adopts Tucker decomposition. 
IVR~\cite{xiao2024ivr} introduces intermediate-variable regularization to reduce overfitting and improve generalization.

\begin{figure*}
    \centering
    \includegraphics[width=0.97\linewidth]{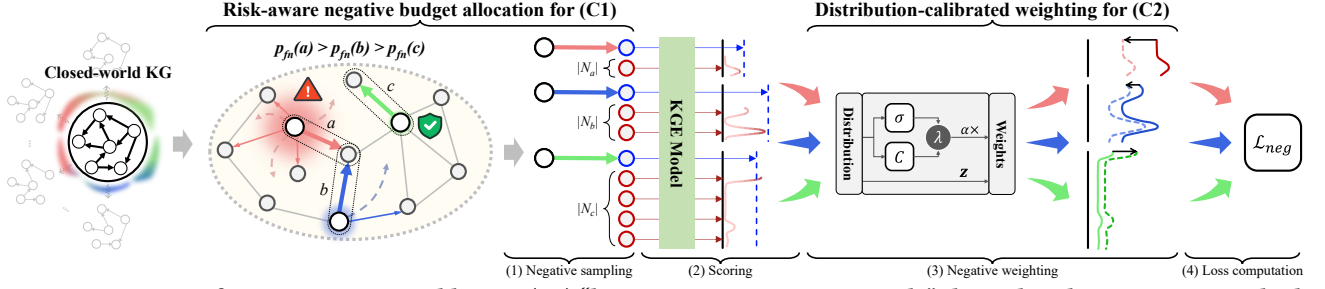}
    \vspace{-2mm}
    \caption{Overview of {\method}. {\method} addresses (C1) ``how many negatives to sample'' through risk-aware negative budget allocation and (C2) ``how much should each negative contribute'' through distribution-calibrated negative weighting.}
    \label{fig:overview}
    \vspace{-3mm}
\end{figure*}

\vspace{1mm}
\noindent
\textbf{Contrastive learning for KGE}.
KGE models are generally trained through contrastive learning with negative samples.
Accordingly, various methods have been proposed to improve negative selection and utilization.
Uniform sampling~\cite{bordes2013translating} randomly corrupts entities, whereas KBGAN~\cite{cai2018kbgan} uses adversarial learning to generate informative negatives.
NSCaching~\cite{zhang2019nscaching} reuses high-scoring negatives, while \sa~\cite{sun2019rotate} assigns larger weights to high-scoring negatives.
SANS~\cite{ahrabian2020sans} exploits local graph structure, whereas EANS~\cite{je2022eans} samples nearby entities in embedding space and mitigates false negatives with an auxiliary loss.
ERDNS~\cite{yao2023erdns}, motivated by optimal embedding analysis~\cite{kamigaito2022comprehensive}, adaptively allocates a fixed batch-level negative budget across positives based on their degrees. 
Unlike {\method}, however, its adaptation is batch-specific and does not account for open-world false-negative risk.
DeMix~\cite{chen2023demix} identifies likely false and true negatives based on model scores and mixes them with positives and negatives, respectively, to mitigate false negatives while generating harder negatives.
M$^2$ixKG~\cite{che2024m2ixkg} synthesizes hard negatives by mixing same-relation or high-scoring entities in embedding space.
IF-NS~\cite{cai2025ifns} prioritizes negatives by their estimated training influence.
BigramKG~\cite{che2025bigramkg} down-weights potential false negatives using relation--entity semantic matching.
ConDNS~\cite{wang2025condns} generates synthetic negatives via conditional diffusion, with difficulty controlled by the denoising timestep.

\section{Preliminary: Contrastive Learning for KGE}\label{sec:preliminary}
Let $\mathcal{G}=(\mathcal{E},\mathcal{R},\mathcal{T})$ denote a KG with entity, relation, and observed triple sets $\mathcal{E}$, $\mathcal{R}$, and $\mathcal{T}$, respectively.
Given $\mathcal{G}$ and a KGE model $\theta$, contrastive learning with negative samples proceeds as follows:
\begin{itemize}[leftmargin=10pt]
    \item \textbf{(1) Negative sampling.} For each positive triple $(h,r,t)\in\mathcal{T}$, negative triples are generated by replacing its head or tail entity. 
    
    \item \textbf{(2) Scoring.} The KGE model computes scores for the positive and its sampled negatives, where a higher score indicates that the triple is more likely to be true.

    \item \textbf{(3) Negative weighting.} Each sampled negative is assigned a weight that determines its contribution to the training objective. 
    
    \item \textbf{(4) Loss computation.} A contrastive loss is computed using the positive and negative scores to update the model $\theta$.
\end{itemize}

In practice, for (1), a fixed number of negatives (e.g., 128 or 256) is typically generated for each positive, while the selection of negatives depends on the sampling strategy.
For (3), negative weight assignment varies across methods: uniform weighting assigns equal weights to all negatives, whereas \sa~\cite{sun2019rotate} assigns different weights based on their model scores.
For (4), the loss function for contrastive learning with negative samples is defined as follows:
\begin{equation}
\mathcal{L}
=
-\frac{1}{2|\mathcal{P}|}
\sum_{(\x,y^+)\in\mathcal{P}}
\left[
\log\sigma(\theta(\x,y^+))
+
\sum_{y^-_i\in\mathcal{N}_{\x,y^+}}
w_{\x,i}\log\sigma(-\theta(\x,y^-_i))
\right],
\label{eq:contrastive-loss}  
\end{equation}
where $\mathcal{P}$ denotes the set of positive triples, $\x=(h,r)$ or $(r,t)$ denotes an anchor for a positive $(h,r,t)$,
$\mathcal{N}_{\x,y^+}$ denotes its set of sampled negative targets, and $\theta(\x,y)$ and $w_{\x,i}$ denote the score of target $y$ and the weight of the $i$-th negative, respectively.
After training, the learned representations of entities and relations can be utilized for various downstream tasks~\cite{mohamed2021biological,song2026kg-chem,wang2019kg-rec,luo2026gfm,yasunaga2022lang-kg}.
In this work, we focus on knowledge graph completion (KGC), which predicts the missing entity in $(h,r,?)$ or $(?,r,t)$. 
For each query, candidate entities are ranked by their scores, and the resulting rankings are evaluated using KGC metrics~\cite{moon2025sharp,sun2020reevaluation,moon2026generalized}.

In this work, we focus on determining \textit{how many negatives to sample} in \textbf{Step (1)} (Section~\ref{sec:proposed-q1-budget}) and \textit{how much each negative should contribute to learning} in \textbf{Step (3)} (Section~\ref{sec:proposed-q2-weight}).

\section{Proposed Framework: {\method}}\label{sec:proposed}
In this section, we present {\method}, a novel contrastive learning framework for KGE.
As illustrated in Figure~\ref{fig:overview}, 
{\method} integrates two strategies: \textbf{(a)} risk-aware negative budget allocation for (C1) and \textbf{(b)} distribution-calibrated negative weighting for (C2).


\subsection{Bridging Open-World and Closed-World KGs}\label{sec:proposed-owa}
To quantify the false-negative risk for each positive, we first analyze the relationship between an observed open-world KG $\mathcal{G}$ and its underlying closed-world counterpart $\mathcal{G}^*$ that contains all latent true facts.
Specifically, we view $\mathcal{G}$ as a \textit{stochastic partial observation} of $\mathcal{G}^*$, where each latent fact is independently observed through a Bernoulli trial,
assuming that $\mathcal{G}$ and $\mathcal{G}^*$ share the same entity and relation sets.
Let $\mathcal{O}\in(0,1]^{|\mathcal{E}|\times|\mathcal{R}|\times|\mathcal{E}|}$ denote an observability tensor, where $\mathcal{O}[h,r,t]$ corresponds to the probability that a latent fact $(h,r,t)\in\mathcal{G}^*$ is observed in $\mathcal{G}$.
Then, the observation process is formulated as follows:
\begin{equation}
\mathcal{G}
=
\mathcal{M}\odot\mathcal{G}^*,
\quad
\mathcal{M}[h,r,t]
\sim
\operatorname{Bernoulli}\!\left(\mathcal{O}[h,r,t]\right),
\label{eq:observability}
\end{equation}
where $\mathcal{M}\in\{0,1\}^{|\mathcal{E}|\times|\mathcal{R}|\times|\mathcal{E}|}$ is the resulting binary observation mask and $\odot$ denotes the element-wise product.

For an anchor $\x$, i.e., $\x=(h,r)$ or $(r,t)$ for a positive triple $(h,r,t)$, 
let $d^*(\x)$ denote the number of target entities $y_i$ such that $(\x,y_i)$ forms a true fact in $\mathcal{G}^*$.
Before the observation process is realized, its observed degree can be represented as a random variable
\begin{equation}
D(\x)
=
\sum_{i=1}^{d^*(\x)} \mathcal{M}_{\x,y_i},
\qquad
\mathcal{M}_{\x,y_i}\sim\operatorname{Bernoulli}(\epsilon_{\x,y_i}),
\label{eq:observed-degree}
\end{equation}
where $\epsilon_{\x,y_i}$ denotes the observability of the corresponding latent fact, i.e., its entry in $\mathcal{O}$.
After realization, $D(\x)$ takes the observed value $d(\x)$ in $\mathcal{G}$.

Since the individual observability $\epsilon_{\x,y_i}$ cannot be determined without knowledge of the underlying closed-world KG,
we assume that all latent facts are observed with a uniform probability, i.e., $\epsilon_{\x,y_i}=\epsilon$, following a conventional assumption adopted in prior studies~\cite{robinson2020hard,yang2022rethinking}.
Under this assumption, the observed-degree random variable and its mean and variance can be written as 
\begin{equation}
D(\x)\sim\operatorname{Binomial}\!\left(d^*(\x),\epsilon\right),
\label{eq:observed-degree}
\end{equation}
\begin{equation}
\mathbb{E}[D(\x)]
=
\epsilon d^*(\x),
\quad
\operatorname{Var}[D(\x)]
=
\epsilon(1-\epsilon)d^*(\x).
\label{eq:observed-degree-mean-variance}
\end{equation}

Based on this relationship between the anchor degrees in the open- and closed-world KGs, 
we can inversely estimate the latent degree from its observed degree as
\begin{equation}
\widehat{d}^*(\x)
=
\min\left\{
\frac{d(\x)}{\epsilon},
|\mathcal{E}|
\right\}.
\label{eq:latent-degree}
\end{equation}

Based on the estimated latent degree, we quantify the false-negative risk for each positive triple through its anchor $\x$.
The risk can be defined as the fraction of unobserved target entities that are actually true in the latent closed-world KG.
Specifically, among the $|\mathcal{E}|-d(\x)$ target entities unobserved for $\x$ in $\mathcal{G}$, $d^*(\x)-d(\x)$ correspond to latent true facts in $\mathcal{G}^*$.
Since $d^*(\x)$ is unobservable, we estimate the false-negative risk using Eq.~(\ref{eq:observed-degree-mean-variance}) and Eq.~(\ref{eq:latent-degree}) as
\begin{equation}
\widehat{p}_{\mathrm{fn}}(\x)
=
\frac{\widehat{d}^*(\x)-d(\x)}
     {|\mathcal{E}|-d(\x)}
\overset{\frac{d(\x)}{\epsilon} \leq |\mathcal{E}|}{=}
\frac{1-\epsilon}{\epsilon}
\frac{d(\x)}
     {|\mathcal{E}|-d(\x)}.
\label{eq:estimated-fn-risk}
\end{equation}


Since $(1-\epsilon)/\epsilon$ is constant across all positives, the estimated false-negative risk is determined solely by $d(\x)/(|\mathcal{E}|-d(\x))$, which monotonically increases with the observed degree $d(\x)$.
Thus, a positive with a higher-degree anchor has a higher false-negative risk, whereas one with a lower-degree anchor has a lower risk.
This result allows us to use the observed anchor degree $d(\x)$ as a direct proxy for false-negative risk.
Based on this analysis, we next develop a risk-aware strategy that determines the negative budget for each positive according to its observed anchor degree.

\subsection{How Many? Risk-Aware Negative Budget Allocation}\label{sec:proposed-q1-budget}
Our analysis in Section~\ref{sec:proposed-owa} implies that the conventional fixed-budget strategy, which assigns the same number of negatives to every positive, overlooks varying false-negative risks across positive triples.
To address this issue, we allocate fewer negatives to positives with higher-risk anchors and more negatives to those with lower-risk anchors.
Specifically, we determine the negative budget for each positive $(\x,y^+)$ based on its anchor degree as
\begin{equation}
N(\x,y^+)
=
\max\{1,\left\lfloor
N_{\max}
\left(
\frac{d_{\min}}{d(\x)}
\right)^{\gamma}\right\rceil\},
\quad
\gamma\geq0,
\label{eq:negative-budget}
\end{equation}
where $N_{\max}$ denotes the maximum negative budget, $d_{\min}=\min_{\x}d(\x)$ is the minimum observed anchor degree, and $\gamma$ controls the sensitivity of the budget to the estimated risk.
An anchor with $d(\x)=d_{\min}$, which has the lowest estimated risk, receives the maximum budget $N_{\max}$, while the budget progressively decreases as $d(\x)$ increases.
When $\gamma=0$, Eq.~(\ref{eq:negative-budget}) reduces to conventional fixed-size negative sampling, whereas a larger $\gamma$ more aggressively reduces the negative budget for higher-risk anchors.
In this way, our strategy mitigates false-negative exposure without explicitly identifying individual false negatives.

We further consider two adjustments to account for additional factors that may affect the observed anchor degree $d(\x)$: (a) observation uncertainty and (b) relation-level degree characteristics.

\vspace{0.5mm}
\noindent
\textbf{(a) Uncertainty-aware adjustment.}
The observed degree $d(\x)$ is a realization of the random variable $D(\x)$ and may thus deviate from its expected value even under uniform observability.
To account for this observation uncertainty, we perturb $d(\x)$ according to its degree-dependent variance:
\begin{equation}
\widetilde{d}_{u}(\x)
=
\max\{1,d(\x)+\eta_{\x}\},
\qquad
\eta_{\x}\sim\mathcal{N}\!\left(0,\tau d(\x)\right),
\label{eq:uncertainty-degree}
\end{equation}
where $\tau\geq0$ controls the degree of uncertainty and absorbs the unknown observability factor in the variance derived in Eq.~(\ref{eq:observed-degree-mean-variance}).
When applying this adjustment, we use $\widetilde{d}_{u}(\x)$ in place of $d(\x)$ in Eq.~(\ref{eq:negative-budget}); $\tau=0$ recovers the original risk-aware allocation.

\vspace{0.5mm}
\noindent
\textbf{(b) Relation-aware adjustment.}
The observed degree $d(\x)$ may also deviate from its expected degree associated with its relation.
We therefore additionally consider relation-level information by smoothing the anchor degree toward the average degree of anchors with the same relation:
\begin{equation}
\widetilde{d}_{r}(\x)
=
(1-\omega)d(\x)+\omega d_r,
\qquad
d_r
=
\mathbb{E}_{\x\in\mathcal{X}_r}[d(\x)]
\label{eq:relation-degree}
\end{equation}
where $\mathcal{X}_r$ denotes the set of anchors associated with relation $r$, and $\omega\in[0,1]$ controls the contribution of relation-level information.
When applying this adjustment, we use $\widetilde{d}_{r}(\x)$ in place of $d(\x)$ in Eq.~(\ref{eq:negative-budget}), where $\omega=0$ recovers the original risk-aware allocation.

\subsection{How Much? Distribution-Calibrated Negative Weighting}\label{sec:proposed-q2-weight}
After determining the negative budget for each positive, 
we now address (i) \textit{how to sample the allocated number of negatives} and (ii) \textit{how much each sampled negative should contribute to learning}.
For (i), we randomly sample the allocated number of negatives since directly identifying hard negatives requires scoring a large number of candidate entities, following {\sa}~\cite{sun2019rotate}.
For (ii), we first analyze the weighting mechanism of {\sa}, which differentiates the contributions of randomly sampled negatives by assigning larger weights to higher-scoring ones as follows:
\begin{equation}
w_{\x,i}
=
\operatorname{softmax}
\left(
\alpha s_{\x,i}
\right)
=
\frac{\exp(\alpha s_{\x,i})}
{\sum_{j\in\mathcal{N}_{\x,y^+}}\exp(\alpha s_{\x,j})},
\label{eq:self-adv}
\end{equation}
where $s_{\x,i}=\theta(\x,y_i^-)$ is the score of the $i$-th negative and $\alpha$ is a global scaling factor.

To better understand its weighting mechanism, we decompose each negative score as
\begin{equation}
s_{\x,i}
=
\mu_\x+\sigma_\x z_{\x,i},
\qquad
z_{\x,i}
=
\frac{s_{\x,i}-\mu_\x}{\sigma_\x},
\end{equation}
where $\mu_\x$ and $\sigma_\x$ denote the mean and standard deviation of the negative scores, respectively, and $z_{\x,i}$ is the standardized relative score of the $i$-th negative.
Since softmax is invariant to a common additive shift, Eq.~(\ref{eq:self-adv}) can be equivalently expressed as
\begin{equation}
w_{\x,i}
=
\operatorname{softmax}
\left(
\alpha (\mu_\x+\sigma_\x z_{\x,i})
\right)
=
\operatorname{softmax}
\left(
\alpha\sigma_\x z_{\x,i}
\right).
\label{eq:self-adv-decomposition}
\end{equation}
This decomposition shows that $z_{\x,i}$ determines the relative score configuration among negatives, whereas $\alpha\sigma_\x$ determines how strongly their differences are emphasized.
Since $\alpha$ is globally shared across anchors, the anchor-specific weighting strength of {\sa} is determined by the score dispersion $\sigma_\x$.

However, dispersion alone may not sufficiently characterize how much each negative should contribute to learning.
As observed for P2 in Figure~\ref{fig:observation}(b), high scores can be strongly concentrated on a few negatives even when their overall score dispersion is relatively small.
In such cases, a small $\sigma_\x$ makes {\sa} weights more uniform, potentially reducing the contributions of these concentrated high-scoring negatives.
This motivates us to additionally consider \textit{score concentration} when determining the weighting strength.

To quantify score concentration, we construct a probability distribution from the standardized negative scores:
\begin{equation}
\pi_{\x,i}
=
\frac{\exp(z_{\x,i})}
{\sum_{j\in\mathcal{N}_{\x,y^+}}\exp(z_{\x,j})}.
\label{eq:score-distribution}
\end{equation}
We then define the concentration of probability mass on high-scoring negatives using normalized entropy as
\begin{equation}
C_\x
=
1-
\frac{H(\pi_\x)}
{\log|\mathcal{N}_{\x,y^+}|},
\qquad
H(\pi_\x)
=
-\sum_{i\in\mathcal{N}_{\x,y^+}}
\pi_{\x,i}\log\pi_{\x,i},
\label{eq:score-concentration}
\end{equation}
where $C_\x\in[0,1]$.
A larger $C_\x$ indicates that high-score probability mass is concentrated on fewer negatives.

Finally, we interpolate between score dispersion and concentration to determine the weighting strength:
\begin{equation}
w_{\x,i}
=
\operatorname{softmax}_i
\left(
\alpha
\left[
(1-\lambda)\sigma_\x+\lambda C_\x
\right]
z_{\x,i}
\right),
\label{eq:distribution-calibrated-weight}
\end{equation}
where $\lambda\in[0,1]$ controls the contribution of score concentration.

\begin{figure}[t]
\centering
\setlength\tabcolsep{0.6pt}
\begin{tabular}{cc}
\includegraphics[width=0.435\linewidth]{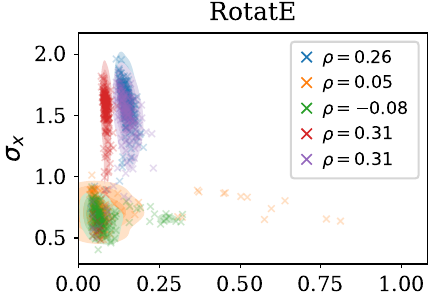} & 
\includegraphics[width=0.435\linewidth]{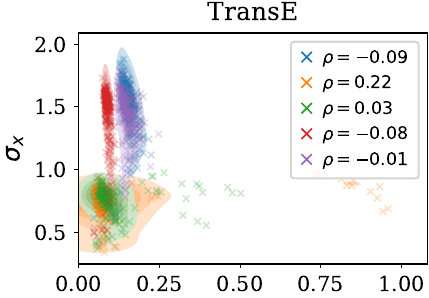} \\
\includegraphics[width=0.435\linewidth]{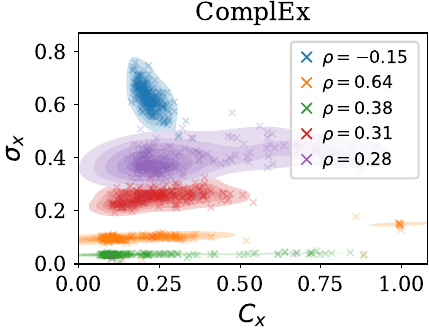} & 
\includegraphics[width=0.435\linewidth]{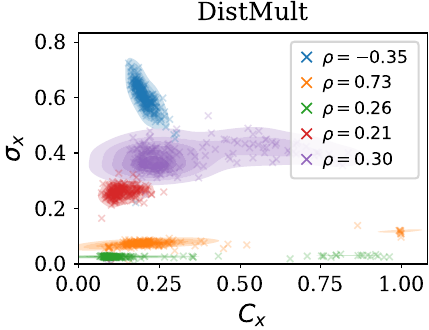} 
\end{tabular}
\vspace{-3mm}
\caption{Relationship between dispersion ($\sigma_\x$) and concentration ($C_\x$) of negative-score distributions during training with {\sa} on FB15k237. Each color represents the evolution of the negative-score distribution for the same positive.}
\label{fig:disp-concentration}
\vspace{-4mm}
\end{figure}

\vspace{0.5mm}
\noindent
\textbf{Analysis of dispersion and concentration.}
To further justify our design, we analyze the relationship between score dispersion $\sigma_\x$ and concentration $C_\x$ during training in Figure~\ref{fig:disp-concentration}, 
where each point represents the negative-score distribution of a positive and each color tracks its evolution during training.
The two characteristics are not strongly correlated (low $\rho$ values) and exhibit distinct behaviors: concentration varies substantially while dispersion remains relatively stable for ComplEx and DistMult, whereas the opposite pattern appears for RotatE and TransE.
These observations suggest that dispersion and concentration provide complementary information, motivating us to consider both in determining the anchor-specific weighting strength.

\vspace{0.5mm}
\noindent
\textbf{Budget-aware weight normalization.}
Unlike conventional weighting with a fixed negative budget, {\method} assigns different numbers of negatives to individual positives.
Since the negative weights for each positive $(\x,y^+)$ are normalized as $\sum_{i\in\mathcal{N}_{\x,y^+}}w_{\x,i}=1$, the average weight per negative becomes $1/N(\x,y^+)$.
Consequently, a high-risk positive assigned a smaller negative budget can have larger weights on its fewer negatives.
To control this effect, we introduce a budget-aware adjustment as follows:
\begin{equation}
\widetilde{w}_{\x,i}
=
\frac{|\mathcal{P}|N(\x,y^+)^\beta}
{\sum_{(\x',y')\in\mathcal{P}}N(\x',y')^\beta}
w_{\x,i},
\qquad
\beta\geq0,
\label{eq:budget-aware-weight}
\end{equation}
where $\beta$ controls the degree of weight adjustment according to the allocated negative budget.
When $\beta=0$, negatives from smaller-budget positives have larger average weights due to the $1/N(\x,y^+)$ effect.
As $\beta$ increases, the weights are adjusted according to the allocated budget: negatives from smaller-budget positives are down-weighted, while those from larger-budget positives are up-weighted.
In particular, when $\beta=1$, the effect of the budget size on the average per-negative weight is fully compensated, resulting in the same average per-negative weight across positives.
Finally, the negative weights $\widetilde{w}_{\x,i}$ computed by Eq.~(\ref{eq:budget-aware-weight}) are applied to the contrastive loss in Eq.~(\ref{eq:contrastive-loss}) to train the KGE model.


\vspace{0.5mm}
\noindent
\textbf{Connection to existing methods.}
{\method} generalizes two existing methods as special cases.
With $\gamma=0$, $\beta=0$, and $\alpha=0$, {\method} reduces to the \textit{fixed-uniform method}, using a fixed negative budget with uniform weighting.
With $\gamma=0$, $\lambda=0$, and $\beta=0$, {\method} reduces to {\sa}, using a fixed negative budget with dispersion-only weighting.
Thus, {\method} provides a unified generalization of existing fixed-budget negative sampling while adaptively controlling both the number and contribution of negative samples.

Overall, {\method} addresses the two fundamental challenges in contrastive learning with negative samples: it determines \textit{how many} negatives to sample for each positive through risk-aware negative budget allocation and \textit{how much} each negative should contribute through distribution-calibrated negative weighting.

\section{Experimental Validation}\label{sec:eval}

In this section, we aim to answer the following evaluation questions:
\begin{itemize}[leftmargin=3mm, labelindent=0pt]
\item \textbf{EQ1} (\textit{KGE Performance}). To what extent does {\method} improve the performance of KGE models compared to existing methods?
\item \textbf{EQ2} (\textit{Training efficiency}). How efficient is the training of {\method} compared to existing methods?
\item \textbf{EQ3} (\textit{Ablation study}). How effective is each of our proposed strategy in improving the performance of {\method}?
\item \textbf{EQ4} (\textit{Sensitivity analysis}). How sensitive is the performance of {\method} to its hyperparameters?
\end{itemize}

\begin{table*}[t]
\centering
\footnotesize
\setlength{\tabcolsep}{7pt}
\def\arraystretch{0.65} 
\caption{KGE performance (MRR and H@10) comparison of {\method} with baselines. {\method} achieves the best average rank on all datasets.
$*$ indicates results reported in the original works. \textcolor{myblue}{Bold} and \textcolor{gray}{\underline{underline}} indicate the best and the second-best results, respectively. A.R. and OOM denote average rank and out of memory, respectively.}
\vspace{-1mm}
\begin{tabular}{cl|cc|cc|cc|cc|c}
\toprule

\multirow{2}{*}{Dataset}
& \multirow{2}{*}{Method}
& \multicolumn{2}{c}{RotatE}
& \multicolumn{2}{c}{TransE}
& \multicolumn{2}{c}{ComplEx}
& \multicolumn{2}{c|}{DistMult}
& \multirow{2}{*}{A.R.$\downarrow$} \\

\cmidrule(lr){3-10}

&
& MRR $\uparrow$ & H@10 $\uparrow$
& MRR $\uparrow$ & H@10 $\uparrow$
& MRR $\uparrow$ & H@10 $\uparrow$
& MRR $\uparrow$ & H@10 $\uparrow$ \\

\midrule

\multirow[c]{16}{*}{FB15k237}

& KBGAN$^*$ \VN{[NAACL'18]} &
- & - &
29.4{\scriptsize\color{gray}{$\pm \text{N/A}$}} & 49.7{\scriptsize\color{gray}{$\pm \text{N/A}$}} &
28.2{\scriptsize\color{gray}{$\pm \text{N/A}$}} & 45.4{\scriptsize\color{gray}{$\pm \text{N/A}$}} &
26.7{\scriptsize\color{gray}{$\pm \text{N/A}$}} & 45.3{\scriptsize\color{gray}{$\pm \text{N/A}$}} & - \\

& NSCaching$^*$ \VN{[ICDE'19]} &
30.9{\scriptsize\color{gray}{$\pm \text{N/A}$}} & 48.6{\scriptsize\color{gray}{$\pm \text{N/A}$}} &
30.0{\scriptsize\color{gray}{$\pm \text{N/A}$}} & 47.4{\scriptsize\color{gray}{$\pm \text{N/A}$}} & 
30.2{\scriptsize\color{gray}{$\pm \text{N/A}$}} & 48.1{\scriptsize\color{gray}{$\pm \text{N/A}$}} &
28.3{\scriptsize\color{gray}{$\pm \text{N/A}$}} & 45.6{\scriptsize\color{gray}{$\pm \text{N/A}$}} & 11.5 \\

& DeMix-Adv$^*$ \VN{[ISWC'23]} &
32.9{\scriptsize\color{gray}{$\pm \text{N/A}$}} & 51.8{\scriptsize\color{gray}{$\pm \text{N/A}$}} &
31.8{\scriptsize\color{gray}{$\pm \text{N/A}$}} & 51.0{\scriptsize\color{gray}{$\pm \text{N/A}$}} &
30.7{\scriptsize\color{gray}{$\pm \text{N/A}$}} & 47.9{\scriptsize\color{gray}{$\pm \text{N/A}$}} &
30.1{\scriptsize\color{gray}{$\pm \text{N/A}$}} & 47.0{\scriptsize\color{gray}{$\pm \text{N/A}$}} & 9.2 \\

& M\textsuperscript{2}ixKG$^*$ \VN{[NN'24]} &
33.7{\scriptsize\color{gray}{$\pm \text{N/A}$}} & 53.2{\scriptsize\color{gray}{$\pm \text{N/A}$}} &
33.1{\scriptsize\color{gray}{$\pm \text{N/A}$}} & 52.9{\scriptsize\color{gray}{$\pm \text{N/A}$}} &
32.3{\scriptsize\color{gray}{$\pm \text{N/A}$}} & 51.6{\scriptsize\color{gray}{$\pm \text{N/A}$}} &
31.7{\scriptsize\color{gray}{$\pm \text{N/A}$}} & 50.3{\scriptsize\color{gray}{$\pm \text{N/A}$}} & 4.1 \\

& IF-NS$^*$ \VN{[KBS'25]} &
- & - &
31.0{\scriptsize\color{gray}{$\pm \text{N/A}$}} & 48.9{\scriptsize\color{gray}{$\pm \text{N/A}$}} &
- & - &
- & - & - \\

& BigramKG$^*$ \VN{[TKDE'25]} &
34.0{\scriptsize\color{gray}{$\pm \text{N/A}$}} & 53.2{\scriptsize\color{gray}{$\pm \text{N/A}$}} &
33.5{\scriptsize\color{gray}{$\pm \text{N/A}$}} & 53.1{\scriptsize\color{gray}{$\pm \text{N/A}$}} &
32.3{\scriptsize\color{gray}{$\pm \text{N/A}$}} & 51.6{\scriptsize\color{gray}{$\pm \text{N/A}$}} &
31.6{\scriptsize\color{gray}{$\pm \text{N/A}$}} & 49.7{\scriptsize\color{gray}{$\pm \text{N/A}$}} & 4.0 \\


& Uniform \VN{[NeurIPS'13]} &
29.5{\scriptsize\color{gray}{$\pm 0.1$}} & 47.9{\scriptsize\color{gray}{$\pm 0.1$}} &
29.3{\scriptsize\color{gray}{$\pm 0.0$}} & 48.1{\scriptsize\color{gray}{$\pm 0.1$}} &
27.2{\scriptsize\color{gray}{$\pm 0.1$}} & 43.2{\scriptsize\color{gray}{$\pm 0.1$}} &
25.5{\scriptsize\color{gray}{$\pm 0.0$}} & 40.4{\scriptsize\color{gray}{$\pm 0.1$}} & 14.6 \\

& Uniform SANS \VN{[EMNLP'20]} &
29.8{\scriptsize\color{gray}{$\pm 0.1$}} & 48.0{\scriptsize\color{gray}{$\pm 0.0$}} &
29.6{\scriptsize\color{gray}{$\pm 0.0$}} & 48.4{\scriptsize\color{gray}{$\pm 0.1$}} &
27.1{\scriptsize\color{gray}{$\pm 0.1$}} & 43.1{\scriptsize\color{gray}{$\pm 0.1$}} &
25.9{\scriptsize\color{gray}{$\pm 0.0$}} & 41.0{\scriptsize\color{gray}{$\pm 0.1$}} & 13.9 \\

& Uniform RW-SANS \VN{[EMNLP'20]} &
29.7{\scriptsize\color{gray}{$\pm 0.0$}} & 48.5{\scriptsize\color{gray}{$\pm 0.0$}} &
31.2{\scriptsize\color{gray}{$\pm 0.1$}} & 50.6{\scriptsize\color{gray}{$\pm 0.0$}} &
29.0{\scriptsize\color{gray}{$\pm 0.1$}} & 46.1{\scriptsize\color{gray}{$\pm 0.2$}} &
27.3{\scriptsize\color{gray}{$\pm 0.1$}} & 43.6{\scriptsize\color{gray}{$\pm 0.1$}} & 11.8 \\

& Self-Adv \VN{[ICLR'19]} &
33.6{\scriptsize\color{gray}{$\pm 0.0$}} & 53.1{\scriptsize\color{gray}{$\pm 0.1$}} &
33.0{\scriptsize\color{gray}{$\pm 0.1$}} & 52.7{\scriptsize\color{gray}{$\pm 0.1$}} &
32.2{\scriptsize\color{gray}{$\pm 0.0$}} & 51.1{\scriptsize\color{gray}{$\pm 0.1$}} &
30.7{\scriptsize\color{gray}{$\pm 0.0$}} & 48.4{\scriptsize\color{gray}{$\pm 0.0$}} & 6.9 \\

& Self-Adv SANS \VN{[EMNLP'20]} &
33.6{\scriptsize\color{gray}{$\pm 0.1$}} & 53.1{\scriptsize\color{gray}{$\pm 0.1$}} &
32.6{\scriptsize\color{gray}{$\pm 0.0$}} & 52.0{\scriptsize\color{gray}{$\pm 0.0$}} &
31.7{\scriptsize\color{gray}{$\pm 0.0$}} & 50.2{\scriptsize\color{gray}{$\pm 0.1$}} &
31.1{\scriptsize\color{gray}{$\pm 0.1$}} & 48.7{\scriptsize\color{gray}{$\pm 0.1$}} & 6.9 \\

& Self-Adv RW-SANS \VN{[EMNLP'20]} &
31.5{\scriptsize\color{gray}{$\pm 0.0$}} & 50.8{\scriptsize\color{gray}{$\pm 0.0$}} &
31.2{\scriptsize\color{gray}{$\pm 0.0$}} & 49.9{\scriptsize\color{gray}{$\pm 0.1$}} &
30.3{\scriptsize\color{gray}{$\pm 0.3$}} & 47.0{\scriptsize\color{gray}{$\pm 0.5$}} &
30.8{\scriptsize\color{gray}{$\pm 0.2$}} & 48.3{\scriptsize\color{gray}{$\pm 0.3$}} & 9.8 \\

& ERDNS \VN{[ISWC'23]} &
\cellcolor{second}\underline{34.5}{\scriptsize\color{gray}{$\pm 0.0$}} & 53.4{\scriptsize\color{gray}{$\pm 0.1$}} &
\cellcolor{second}\underline{34.1}{\scriptsize\color{gray}{$\pm 0.0$}} & \cellcolor{second}\underline{53.3}{\scriptsize\color{gray}{$\pm 0.0$}} &
\cellcolor{second}\underline{35.6}{\scriptsize\color{gray}{$\pm 0.1$}} & \cellcolor{first}\textbf{54.6}{\scriptsize\color{gray}{$\pm 0.1$}} &
\cellcolor{second}\underline{35.0}{\scriptsize\color{gray}{$\pm 0.1$}} & \cellcolor{second}\underline{53.8}{\scriptsize\color{gray}{$\pm 0.1$}} & \cellcolor{second}\underline{2.2} \\

& EANS \VN{[arXiv'22]} &
\cellcolor{second}\underline{34.5}{\scriptsize\color{gray}{$\pm 0.0$}} & \cellcolor{second}\underline{53.7}{\scriptsize\color{gray}{$\pm 0.1$}} &
\cellcolor{second}\underline{34.1}{\scriptsize\color{gray}{$\pm 0.0$}} & \cellcolor{first}\textbf{53.4}{\scriptsize\color{gray}{$\pm 0.0$}} &
29.6{\scriptsize\color{gray}{$\pm 0.1$}} & 48.3{\scriptsize\color{gray}{$\pm 0.6$}} &
30.9{\scriptsize\color{gray}{$\pm 0.2$}} & 49.5{\scriptsize\color{gray}{$\pm 0.2$}} & 4.8 \\

& ConDNS \VN{[NC'25]} &
33.6{\scriptsize\color{gray}{$\pm 0.1$}} & 53.1{\scriptsize\color{gray}{$\pm 0.1$}} &
32.9{\scriptsize\color{gray}{$\pm 0.1$}} & 52.7{\scriptsize\color{gray}{$\pm 0.1$}} &
32.3{\scriptsize\color{gray}{$\pm 0.1$}} & 51.1{\scriptsize\color{gray}{$\pm 0.1$}} &
30.8{\scriptsize\color{gray}{$\pm 0.0$}} & 48.5{\scriptsize\color{gray}{$\pm 0.1$}} & 6.4 \\

& \textbf{\method\text{ }(Ours)} &
\cellcolor{first}\textbf{35.1}{\scriptsize\color{gray}{$\pm 0.1$}} & \cellcolor{first}\textbf{53.9}{\scriptsize\color{gray}{$\pm 0.1$}} &
\cellcolor{first}\textbf{34.7}{\scriptsize\color{gray}{$\pm 0.0$}} & \cellcolor{first}\textbf{53.4}{\scriptsize\color{gray}{$\pm 0.1$}} &
\cellcolor{first}\textbf{35.7}{\scriptsize\color{gray}{$\pm 0.1$}} & \cellcolor{second}\underline{54.4}{\scriptsize\color{gray}{$\pm 0.1$}} &
\cellcolor{first}\textbf{35.4}{\scriptsize\color{gray}{$\pm 0.1$}} & \cellcolor{first}\textbf{54.0}{\scriptsize\color{gray}{$\pm 0.1$}} & \cellcolor{first}\textbf{1.2} \\

\midrule

\multirow[c]{16}{*}{WN18RR}
& KBGAN$^*$ \VN{[NAACL'18]} &
- & - &
18.6{\scriptsize\color{gray}{$\pm \text{N/A}$}} & 45.4{\scriptsize\color{gray}{$\pm \text{N/A}$}} &
42.9{\scriptsize\color{gray}{$\pm \text{N/A}$}} & 47.0{\scriptsize\color{gray}{$\pm \text{N/A}$}} &
38.5{\scriptsize\color{gray}{$\pm \text{N/A}$}} & 44.3{\scriptsize\color{gray}{$\pm \text{N/A}$}} & - \\

& NSCaching$^*$ \VN{[ICDE'19]} &
47.1{\scriptsize\color{gray}{$\pm \text{N/A}$}} & 55.3{\scriptsize\color{gray}{$\pm \text{N/A}$}} &
20.0{\scriptsize\color{gray}{$\pm \text{N/A}$}} & 47.8{\scriptsize\color{gray}{$\pm \text{N/A}$}} &
44.6{\scriptsize\color{gray}{$\pm \text{N/A}$}} & 50.9{\scriptsize\color{gray}{$\pm \text{N/A}$}} &
41.3{\scriptsize\color{gray}{$\pm \text{N/A}$}} & 45.5{\scriptsize\color{gray}{$\pm \text{N/A}$}} & 12.6 \\

& DeMix-Adv$^*$ \VN{[ISWC'23]} &
47.9{\scriptsize\color{gray}{$\pm \text{N/A}$}} & 57.6{\scriptsize\color{gray}{$\pm \text{N/A}$}} &
22.0{\scriptsize\color{gray}{$\pm \text{N/A}$}} & 52.1{\scriptsize\color{gray}{$\pm \text{N/A}$}} &
46.8{\scriptsize\color{gray}{$\pm \text{N/A}$}} & 55.2{\scriptsize\color{gray}{$\pm \text{N/A}$}} &
43.9{\scriptsize\color{gray}{$\pm \text{N/A}$}} & 53.5{\scriptsize\color{gray}{$\pm \text{N/A}$}} & 6.6  \\

& M\textsuperscript{2}ixKG$^*$ \VN{[NN'24]} &
47.7{\scriptsize\color{gray}{$\pm \text{N/A}$}} & 57.5{\scriptsize\color{gray}{$\pm \text{N/A}$}} &
23.0{\scriptsize\color{gray}{$\pm \text{N/A}$}} & \cellcolor{second}\underline{54.1}{\scriptsize\color{gray}{$\pm \text{N/A}$}} &
47.4{\scriptsize\color{gray}{$\pm \text{N/A}$}} & 56.2{\scriptsize\color{gray}{$\pm \text{N/A}$}} &
44.6{\scriptsize\color{gray}{$\pm \text{N/A}$}} & 54.1{\scriptsize\color{gray}{$\pm \text{N/A}$}} & \cellcolor{second}\underline{4.1}  \\

& IF-NS$^*$ \VN{[KBS'25]} &
- & - &
23.6{\scriptsize\color{gray}{$\pm \text{N/A}$}} & 50.9{\scriptsize\color{gray}{$\pm \text{N/A}$}} &
- & - &
- & - & -  \\

& BigramKG$^*$ \VN{[TKDE'25]} &
47.7{\scriptsize\color{gray}{$\pm \text{N/A}$}} & 57.3{\scriptsize\color{gray}{$\pm \text{N/A}$}} &
22.3{\scriptsize\color{gray}{$\pm \text{N/A}$}} & 52.9{\scriptsize\color{gray}{$\pm \text{N/A}$}} &
47.3{\scriptsize\color{gray}{$\pm \text{N/A}$}} & \cellcolor{second}\underline{56.8}{\scriptsize\color{gray}{$\pm \text{N/A}$}} &
44.3{\scriptsize\color{gray}{$\pm \text{N/A}$}} & 54.3{\scriptsize\color{gray}{$\pm \text{N/A}$}} & 5.7 \\


& Uniform \VN{[NeurIPS'13]} &
47.5{\scriptsize\color{gray}{$\pm 0.1$}} & 55.7{\scriptsize\color{gray}{$\pm 0.4$}} &
20.5{\scriptsize\color{gray}{$\pm 0.0$}} & 50.0{\scriptsize\color{gray}{$\pm 0.1$}} &
46.3{\scriptsize\color{gray}{$\pm 0.2$}} & 53.8{\scriptsize\color{gray}{$\pm 0.2$}} &
42.8{\scriptsize\color{gray}{$\pm 0.2$}} & 52.6{\scriptsize\color{gray}{$\pm 0.1$}} & 10.8 \\

& Uniform SANS \VN{[EMNLP'20]} &
47.8{\scriptsize\color{gray}{$\pm 0.0$}} & 56.0{\scriptsize\color{gray}{$\pm 0.2$}} &
20.5{\scriptsize\color{gray}{$\pm 0.0$}} & 50.0{\scriptsize\color{gray}{$\pm 0.1$}} &
42.7{\scriptsize\color{gray}{$\pm 0.2$}} & 47.6{\scriptsize\color{gray}{$\pm 0.3$}} &
41.2{\scriptsize\color{gray}{$\pm 0.3$}} & 47.7{\scriptsize\color{gray}{$\pm 1.1$}} & 11.4 \\

& Uniform RW-SANS \VN{[EMNLP'20]} &
48.0{\scriptsize\color{gray}{$\pm 0.1$}} & 56.8{\scriptsize\color{gray}{$\pm 0.2$}} &
21.9{\scriptsize\color{gray}{$\pm 0.0$}} & 51.0{\scriptsize\color{gray}{$\pm 0.2$}} &
40.7{\scriptsize\color{gray}{$\pm 0.3$}} & 43.6{\scriptsize\color{gray}{$\pm 0.2$}} &
41.8{\scriptsize\color{gray}{$\pm 0.2$}} & 49.9{\scriptsize\color{gray}{$\pm 0.2$}} & 10.0 \\

& Self-Adv \VN{[ICLR'19]} &
47.7{\scriptsize\color{gray}{$\pm 0.1$}} & 56.3{\scriptsize\color{gray}{$\pm 0.5$}} &
22.5{\scriptsize\color{gray}{$\pm 0.1$}} & 53.2{\scriptsize\color{gray}{$\pm 0.1$}} &
46.8{\scriptsize\color{gray}{$\pm 0.1$}} & 55.5{\scriptsize\color{gray}{$\pm 0.1$}} &
43.9{\scriptsize\color{gray}{$\pm 0.1$}} & 53.3{\scriptsize\color{gray}{$\pm 0.1$}} & 6.9 \\

& Self-Adv SANS \VN{[EMNLP'20]} &
47.7{\scriptsize\color{gray}{$\pm 0.1$}} & 56.0{\scriptsize\color{gray}{$\pm 0.2$}} &
22.5{\scriptsize\color{gray}{$\pm 0.1$}} & 53.1{\scriptsize\color{gray}{$\pm 0.1$}} &
37.5{\scriptsize\color{gray}{$\pm 0.4$}} & 40.0{\scriptsize\color{gray}{$\pm 0.5$}} &
38.2{\scriptsize\color{gray}{$\pm 0.4$}} & 42.6{\scriptsize\color{gray}{$\pm 0.5$}} & 11.6 \\

& Self-Adv RW-SANS \VN{[EMNLP'20]} &
48.2{\scriptsize\color{gray}{$\pm 0.1$}} & 57.2{\scriptsize\color{gray}{$\pm 0.2$}} &
22.3{\scriptsize\color{gray}{$\pm 0.3$}} & 51.1{\scriptsize\color{gray}{$\pm 0.4$}} &
23.0{\scriptsize\color{gray}{$\pm 0.2$}} & 34.4{\scriptsize\color{gray}{$\pm 0.1$}} &
40.0{\scriptsize\color{gray}{$\pm 0.4$}} & 45.1{\scriptsize\color{gray}{$\pm 0.3$}} & 10.6 \\

& ERDNS \VN{[ISWC'23]} &
47.7{\scriptsize\color{gray}{$\pm 0.1$}} & 56.5{\scriptsize\color{gray}{$\pm 0.4$}} &
\cellcolor{second}\underline{23.9}{\scriptsize\color{gray}{$\pm 0.0$}} & 53.2{\scriptsize\color{gray}{$\pm 0.1$}} &
\cellcolor{second}\underline{48.4}{\scriptsize\color{gray}{$\pm 0.1$}} & 55.7{\scriptsize\color{gray}{$\pm 0.2$}} &
\cellcolor{second}\underline{45.5}{\scriptsize\color{gray}{$\pm 0.1$}} & \cellcolor{second}\underline{54.8}{\scriptsize\color{gray}{$\pm 0.3$}} & 4.4 \\

& EANS \VN{[arXiv'22]} &
\cellcolor{second}\underline{48.8}{\scriptsize\color{gray}{$\pm 0.0$}} & \cellcolor{second}\underline{57.7}{\scriptsize\color{gray}{$\pm 0.1$}} &
22.9{\scriptsize\color{gray}{$\pm 0.0$}} & 53.4{\scriptsize\color{gray}{$\pm 0.1$}} &
45.6{\scriptsize\color{gray}{$\pm 0.1$}} & 53.4{\scriptsize\color{gray}{$\pm 0.1$}} &
43.1{\scriptsize\color{gray}{$\pm 0.1$}} & 51.8{\scriptsize\color{gray}{$\pm 0.1$}} & 5.9 \\

& ConDNS \VN{[NC'25]} &
47.8{\scriptsize\color{gray}{$\pm 0.1$}} & 56.3{\scriptsize\color{gray}{$\pm 0.6$}} &
22.5{\scriptsize\color{gray}{$\pm 0.0$}} & 53.1{\scriptsize\color{gray}{$\pm 0.1$}} &
46.9{\scriptsize\color{gray}{$\pm 0.1$}} & 55.8{\scriptsize\color{gray}{$\pm 0.1$}} &
43.9{\scriptsize\color{gray}{$\pm 0.1$}} & 53.2{\scriptsize\color{gray}{$\pm 0.3$}} & 6.4 \\

& \textbf{\method\text{ }(Ours)} &
\cellcolor{first}\textbf{49.0}{\scriptsize\color{gray}{$\pm 0.2$}} & \cellcolor{first}\textbf{58.2}{\scriptsize\color{gray}{$\pm 0.2$}} &
\cellcolor{first}\textbf{25.9}{\scriptsize\color{gray}{$\pm 0.0$}} & \cellcolor{first}\textbf{55.1}{\scriptsize\color{gray}{$\pm 0.1$}} &
\cellcolor{first}\textbf{49.4}{\scriptsize\color{gray}{$\pm 0.1$}} & \cellcolor{first}\textbf{57.4}{\scriptsize\color{gray}{$\pm 0.1$}} &
\cellcolor{first}\textbf{46.0}{\scriptsize\color{gray}{$\pm 0.1$}} & \cellcolor{first}\textbf{54.9}{\scriptsize\color{gray}{$\pm 0.1$}} & \cellcolor{first}\textbf{1.0} \\

\midrule

\multirow[c]{10}{*}{YAGO3-10}

& Uniform \VN{[NeurIPS'13]} &
44.7{\scriptsize\color{gray}{$\pm 0.1$}} & 64.7{\scriptsize\color{gray}{$\pm 0.1$}} &
41.2{\scriptsize\color{gray}{$\pm 0.1$}} & 63.8{\scriptsize\color{gray}{$\pm 0.1$}} &
39.6{\scriptsize\color{gray}{$\pm 0.3$}} & 54.5{\scriptsize\color{gray}{$\pm 0.2$}} &
36.3{\scriptsize\color{gray}{$\pm 0.2$}} & 49.7{\scriptsize\color{gray}{$\pm 0.2$}} & 6.2 \\

& Uniform SANS \VN{[EMNLP'20]} &
OOM & OOM &
OOM & OOM &
OOM & OOM &
OOM & OOM & - \\

& Uniform RW-SANS \VN{[EMNLP'20]} &
42.0{\scriptsize\color{gray}{$\pm 0.1$}} & 60.1{\scriptsize\color{gray}{$\pm 0.2$}} &
40.1{\scriptsize\color{gray}{$\pm 0.1$}} & 59.4{\scriptsize\color{gray}{$\pm 0.2$}} &
19.4{\scriptsize\color{gray}{$\pm 1.4$}} & 32.7{\scriptsize\color{gray}{$\pm 0.9$}} &
7.1{\scriptsize\color{gray}{$\pm 0.3$}} & 18.7{\scriptsize\color{gray}{$\pm 0.9$}} & 7.6 \\

& Self-Adv \VN{[ICLR'19]} &
\cellcolor{second}\underline{49.6}{\scriptsize\color{gray}{$\pm 0.1$}} & 67.2{\scriptsize\color{gray}{$\pm 0.1$}} &
50.3{\scriptsize\color{gray}{$\pm 0.1$}} & \cellcolor{first}\textbf{67.9}{\scriptsize\color{gray}{$\pm 0.2$}} &
\cellcolor{second}\underline{48.5}{\scriptsize\color{gray}{$\pm 0.2$}} & 66.2{\scriptsize\color{gray}{$\pm 0.2$}} &
46.6{\scriptsize\color{gray}{$\pm 0.1$}} & 64.5{\scriptsize\color{gray}{$\pm 0.1$}} & \cellcolor{second}\underline{2.7} \\

& Self-Adv SANS \VN{[EMNLP'20]} &
OOM & OOM &
OOM & OOM &
OOM & OOM &
OOM & OOM & - \\

& Self-Adv RW-SANS \VN{[EMNLP'20]} &
41.8{\scriptsize\color{gray}{$\pm 0.2$}} & 58.3{\scriptsize\color{gray}{$\pm 0.3$}} &
41.1{\scriptsize\color{gray}{$\pm 0.0$}} & 57.6{\scriptsize\color{gray}{$\pm 0.2$}} &
21.2{\scriptsize\color{gray}{$\pm 1.1$}} & 34.1{\scriptsize\color{gray}{$\pm 0.6$}} &
38.0{\scriptsize\color{gray}{$\pm 0.3$}} & 50.5{\scriptsize\color{gray}{$\pm 0.2$}} & 7.1 \\

& ERDNS \VN{[ISWC'23]} &
48.8{\scriptsize\color{gray}{$\pm 0.1$}} & 66.8{\scriptsize\color{gray}{$\pm 0.2$}} &
49.9{\scriptsize\color{gray}{$\pm 0.1$}} & 67.5{\scriptsize\color{gray}{$\pm 0.2$}} &
\cellcolor{first}\textbf{54.5}{\scriptsize\color{gray}{$\pm 0.2$}} & \cellcolor{second}\underline{70.2}{\scriptsize\color{gray}{$\pm 0.1$}} &
\cellcolor{second}\underline{52.8}{\scriptsize\color{gray}{$\pm 0.1$}} & \cellcolor{second}\underline{69.6}{\scriptsize\color{gray}{$\pm 0.1$}} & 3.1 \\

& EANS \VN{[arXiv'22]} &
47.4{\scriptsize\color{gray}{$\pm 0.1$}} & 66.5{\scriptsize\color{gray}{$\pm 0.2$}} &
48.7{\scriptsize\color{gray}{$\pm 0.1$}} & \cellcolor{second}\underline{67.7}{\scriptsize\color{gray}{$\pm 0.3$}} &
42.3{\scriptsize\color{gray}{$\pm 0.6$}} & 59.6{\scriptsize\color{gray}{$\pm 0.8$}} &
42.3{\scriptsize\color{gray}{$\pm 0.2$}} & 60.4{\scriptsize\color{gray}{$\pm 0.3$}} & 4.8 \\

& ConDNS \VN{[NC'25]} &
49.5{\scriptsize\color{gray}{$\pm 0.2$}} & \cellcolor{second}\underline{67.4}{\scriptsize\color{gray}{$\pm 0.2$}} &
\cellcolor{second}\underline{50.4}{\scriptsize\color{gray}{$\pm 0.1$}} & \cellcolor{first}\textbf{67.9}{\scriptsize\color{gray}{$\pm 0.2$}} &
48.4{\scriptsize\color{gray}{$\pm 0.3$}} & 66.1{\scriptsize\color{gray}{$\pm 0.2$}} &
46.5{\scriptsize\color{gray}{$\pm 0.3$}} & 64.4{\scriptsize\color{gray}{$\pm 0.1$}} & 3.1 \\

& \textbf{\method\text{ }(Ours)} &
\cellcolor{first}\textbf{50.8}{\scriptsize\color{gray}{$\pm 0.2$}} & \cellcolor{first}\textbf{68.0}{\scriptsize\color{gray}{$\pm 0.2$}} &
\cellcolor{first}\textbf{51.8}{\scriptsize\color{gray}{$\pm 0.1$}} & \cellcolor{second}\underline{67.7}{\scriptsize\color{gray}{$\pm 0.1$}} &
\cellcolor{first}\textbf{54.5}{\scriptsize\color{gray}{$\pm 0.2$}} & \cellcolor{first}\textbf{70.3}{\scriptsize\color{gray}{$\pm 0.3$}} &
\cellcolor{first}\textbf{53.0}{\scriptsize\color{gray}{$\pm 0.2$}} & \cellcolor{first}\textbf{69.7}{\scriptsize\color{gray}{$\pm 0.1$}} & \cellcolor{first}\textbf{1.4} \\

\bottomrule
\end{tabular}
\vspace{-3mm}
\label{tab:rq1-accuracy}
\end{table*}

\subsection{Experimental Setup}\label{sec:eval-setup}
\noindent
\textbf{Datasets and KGE models.} We evaluate {\method} with four KGE models, TransE~\cite{bordes2013translating}, RotatE~\cite{sun2019rotate}, DistMult~\cite{yang2015embeddingentitiesrelationslearning}, and ComplEx~\cite{trouillon2016complex},
on three widely used KGs, FB15k237~\cite{toutanova2015observed}, WN18RR~\cite{dettmers2018convolutional}, and YAGO3-10~\cite{dettmers2018convolutional}. 
Since {\method} is model-agnostic, it can be applied to any KGE models without modifying their underlying architectures.
A detailed description of the datasets is provided in Appendix~\ref{app:datasets}.

\vspace{0.5mm}
\noindent
\textbf{Baselines.} 
We compare {\method} with 15 contrastive learning methods that leverage negative samples: Uniform~\cite{bordes2013translating}, KBGAN~\cite{cai2018kbgan}, {\sa}~\cite{sun2019rotate}, NSCaching~\cite{zhang2019nscaching}, Uniform SANS~\cite{ahrabian2020sans}, Uniform RW-SANS~\cite{ahrabian2020sans}, {\sa} SANS~\cite{ahrabian2020sans}, {\sa} RW-SANS~\cite{ahrabian2020sans}, EANS~\cite{je2022eans}, ERDNS~\cite{yao2023erdns}, DeMix-Adv~\cite{chen2023demix}, M$^2$ixKG~\cite{che2024m2ixkg}, IF-NS~\cite{cai2025ifns}, BigramKG~\cite{che2025bigramkg}, and ConDNS~\cite{wang2025condns}.
We use the official implementations for {\sa}, SANS, EANS, ERDNS, and ConDNS. 
For the remaining six methods, we use their reported results (marked $^*$) since their code is unavailable or does not reproduce the reported performance; these results are unavailable for YAGO3-10.
Baseline details are provided in Appendix~\ref{app:baselines}.

\vspace{0.5mm}
\noindent
\textbf{Evaluation protocol.} Following~\cite{sun2019rotate,sun2020reevaluation}, we use two metrics: mean reciprocal rank (MRR) and Hits@10 (H@10), reported as percentages (\%).
We adopt the filtered setting~\cite{bordes2013translating} and tie-breaking protocol~\cite{sun2020reevaluation}. 
We run all experiments with five different random seeds and report the mean performance with standard deviation.
Implementation details are provided in Appendix~\ref{app:implementation-details}.

\subsection{EQ1. KGE Performance}\label{sec:eval-eq1}
\textbf{Overall performance.}
Table~\ref{tab:rq1-accuracy} shows that {\method} consistently improves the performance of four KGE models across all three KGs.
Compared with Uniform, {\method} improves both MRR and H@10 in every model--KG combination, demonstrating the benefit of addressing the two challenges, i.e., negative budget allocation and importance estimation, is beneficial to improving KGE performance.
Moreover, {\method} achieves the best average rank (A.R.) on all three KGs, verifying consistent advantages over all baseline methods.
Notably, on WN18RR, {\method} ranks first on all eight model--metric combinations, with an A.R. of 1.0 compared with 4.1 for the second-best method.
These results demonstrate the importance of the challenges identified in this work and the effectiveness of {\method} in addressing them.

\begin{figure}
\centering
\methodlegend
\setlength{\tabcolsep}{2pt}
\begin{tabular}{cc}
\includegraphics[width=0.45\linewidth]{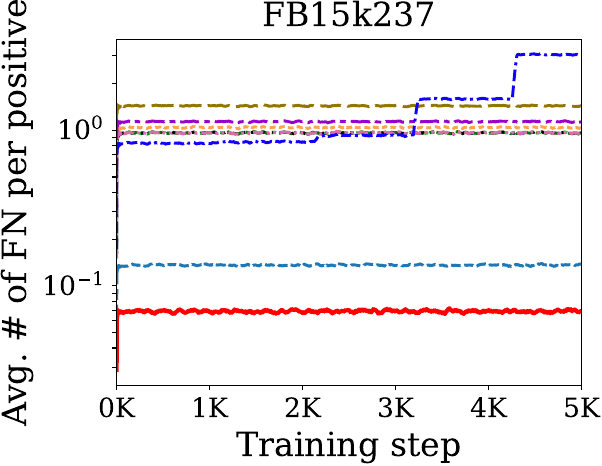} &
\includegraphics[width=0.45\linewidth]{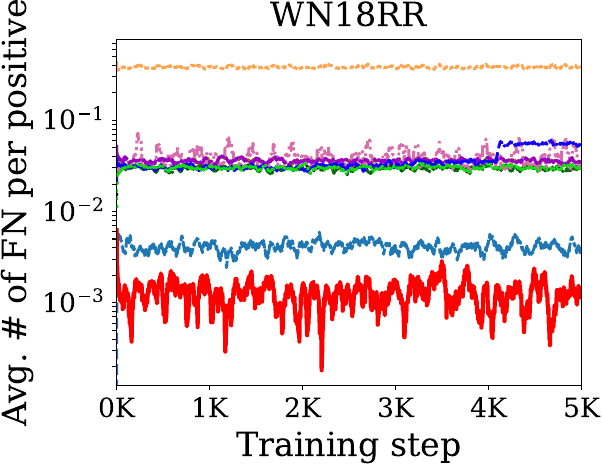}
\end{tabular}
\vspace{-3mm}
\caption{The average number of false negatives sampled per positive during training ($N_{max}=256$). {\method} consistently samples the fewest false negatives among all baselines.}
\label{fig:fn-count}
\vspace{-4mm}
\end{figure}

\vspace{0.5mm}
\noindent
\textbf{Analysis of false-negative sampling risk.}
To further analyze the performance improvement, we examine how many false negatives (FNs) are sampled by each method during training.
Figure~\ref{fig:fn-count} shows the average number of sampled FNs per positive during training on FB15k237 and WN18RR, where a sampled negative is regarded as an FN if it appears as a positive in the validation or test set.
{\method} consistently samples far fewer FNs than all baselines on both KGs, providing empirical support for our observed-degree-based measure of false-negative risk.
Together with the performance gains in Table~\ref{tab:rq1-accuracy}, this result suggests that our risk-aware budget allocation strategy effectively mitigates FN sampling, thereby contributing to improved KGE performance.

\begin{figure}
    \centering
    \includegraphics[width=0.95\linewidth]{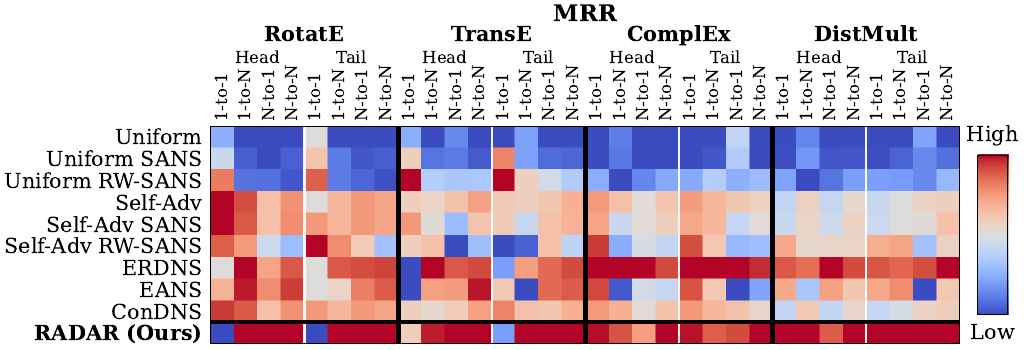}
    \vspace{-1mm}
    \caption{KGE performance across relation cardinality categories for head and tail prediction. {\method} performs strongly on multi-valued relations.}
    \label{fig:rel-cardinality-mrr}
    \vspace{-4mm}
\end{figure}

\vspace{0.5mm}
\noindent
\textbf{Performance by relation cardinality.}
We further analyze the performance (MRR) of each method across 1-to-1, 1-to-N, N-to-1, and N-to-N relation cardinality categories, following~\cite{wang2014transh}.
As shown in Figure~\ref{fig:rel-cardinality-mrr}, KGE models with {\method} outperform the baselines across 1-to-N, N-to-1, and N-to-N relations, where multiple true targets increase the risk of sampling FNs, supporting the effectiveness of our risk-aware budget allocation for such multi-valued relations.
The improvement is less consistent for 1-to-1 relations, which have lower FN risk and account for the fewest triples among the four relation categories (1.6\% of the total), thus having a smaller impact on overall performance.
Detailed statistics and results for all datasets and KGE models are provided in Appendix~\ref{app:relation-category}.

\begin{figure}[b]
    \centering
    \methodlegend
    
    \setlength{\tabcolsep}{1pt}
    \def\arraystretch{0.2} 
    \begin{tabular}{cc}
    \includegraphics[width=0.435\linewidth]{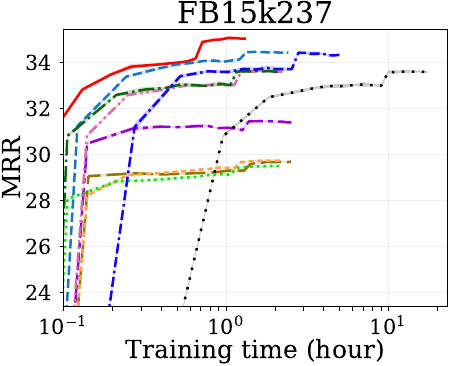} &
    \includegraphics[width=0.435\linewidth]{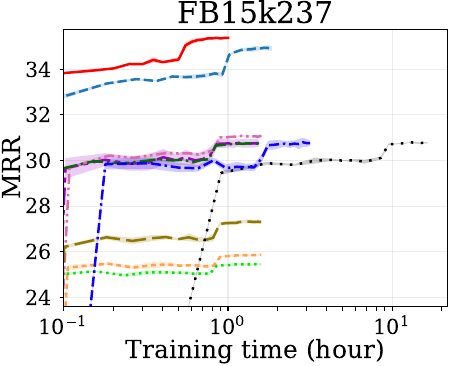} \\
    \includegraphics[width=0.435\linewidth]{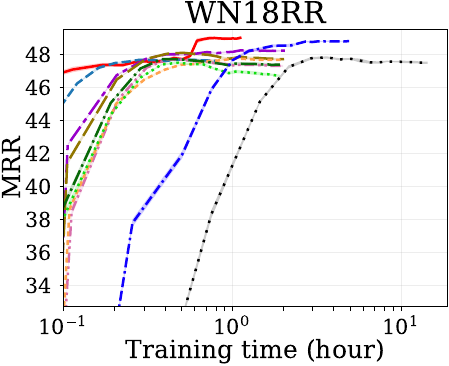} &
    \includegraphics[width=0.435\linewidth]{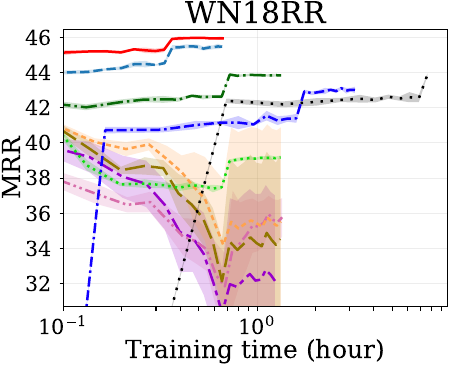} \\
    (a) RotatE & (b) DistMult 
    \end{tabular}
    \vspace{-1mm}
    \caption{Performance--efficiency trade-off of RotatE and DistMult with each method on two KGs. {\method} achieves the highest performance in the shortest training time.}
    \label{fig:rq2-time-performance}
    \vspace{-4mm}
\end{figure}

\subsection{EQ2. Training Efficiency}\label{sec:eval-eq2}
\textbf{Performance--efficiency trade-off.}
In this experiment, we evaluate the training efficiency of {\method} by comparing performance (MRR) against training time.
As shown in Figure~\ref{fig:rq2-time-performance}, {\method} achieves the highest performance while reaching its peak performance faster than all baselines across both KGE models and datasets.
This indicates that our risk-aware negative budget allocation does not simply accelerate training by reducing the number of negatives, but also retains sufficient contrastive supervision for effective KGE training.
By adaptively allocating negative budgets, {\method} reduces the processing of potentially adverse negatives (i.e., high-risk negatives) while preserving effective learning signals, thereby achieving a favorable accuracy--efficiency trade-off.
Results for all KGE models and datasets are provided in Appendix~\ref{app:training-efficiency}.

\vspace{0.5mm}
\noindent
\textbf{Analysis of training resources.}
To further analyze the efficiency gains, we measure training time and GPU memory cost of RotatE on FB15k237.
For a fair comparison, we use the same number of positives $|\mathcal{P}|=1024$, maximum negative budget $N_{max}=256$, and embedding dimension of 1000.
As shown in Table~\ref{tab:rq2-controlled-time-memory}, {\method} reduces training time and memory cost by 30.0\% and 34.2\%, respectively, compared with Uniform and {\sa}, while achieving the highest MRR.
Moreover, unlike three baselines that incur additional computational and memory overhead, {\method} achieves these reductions through its adaptive negative budgets without introducing costly auxiliary procedures.
These results verify that our risk-aware negative budget allocation improves not only KGE performance but also training efficiency.

\begin{table}[t]
\centering
\footnotesize
\caption{Training time (TT) and GPU memory cost (MC) of RotatE with each method on FB15k237.}
\label{tab:rq2-controlled-time-memory}
\setlength{\tabcolsep}{6pt}
\def\arraystretch{0.9}
\begin{tabular}{l|ll|ll|c}
\toprule
Method
& TT $\downarrow$ & Gain
& MC $\downarrow$ & Gain & MRR $\uparrow$ \\
\midrule
Uniform & 1h 40m & \textcolor{gaingrey}{0.0\%} & 7.5GB & \textcolor{gaingrey}{0.0\%} & 29.5 \\
Uniform SANS & 1h 40m & \textcolor{gaingrey}{0.0\%} & 7.5GB & \textcolor{gaingrey}{0.0\%} & 29.8 \\
Uniform RW-SANS & 1h 40m & \textcolor{gaingrey}{0.0\%} & 7.5GB & \textcolor{gaingrey}{0.0\%} & 29.7 \\
Self-Adv & 1h 40m & \textcolor{gaingrey}{0.0\%} & 7.5GB & \textcolor{gaingrey}{0.0\%} & 33.6 \\
Self-Adv SANS & 1h 40m & \textcolor{gaingrey}{0.0\%} & 7.5GB & \textcolor{gaingrey}{0.0\%} & 33.6 \\
Self-Adv RW-SANS & 1h 40m & \textcolor{gaingrey}{0.0\%} & 7.5GB & \textcolor{gaingrey}{0.0\%} & 31.5 \\
ERDNS & 1h 55m & \textcolor{gainred}{+15.0\%} & 7.5GB & \textcolor{gaingrey}{0.0\%} & 34.5 \\
EANS & 4h 25m & \textcolor{gainred}{+165.0\%} & 13.4GB & \textcolor{gainred}{+78.2\%} & 34.5 \\
ConDNS & 15h 50m & \textcolor{gainred}{+850.0\%} & 7.6GB & \textcolor{gainred}{+0.5\%} & 33.6 \\
\midrule
\textbf{\method} & 1h 10m & \textcolor{gainblue}{-30.0\%} & 5.0GB & \textcolor{gainblue}{-34.2\%} & \textbf{35.0} \\

\bottomrule
\end{tabular}
\end{table}

\begin{table}[b]
\caption{Ablation study of {\method}. Ablating either of our strategies consistently degrades KGE performance.}
\label{tab:ablation}
\centering
\footnotesize
\setlength\tabcolsep{4pt}
\def\arraystretch{0.95}
\begin{tabular}{llcccc}
\toprule
Dataset & Method & RotatE & TransE & ComplEx & DistMult \\
\midrule
\multirow{3}{*}{FB15k237}
& {\method} & \textbf{35.1} & \textbf{34.7} & \textbf{35.7} & \textbf{35.4} \\
\cmidrule(lr){2-6}
& w/o RA & 34.1 {\scriptsize\textcolor{gainred}{(-2.7\%)}} & 33.7 {\scriptsize\textcolor{gainred}{(-2.7\%)}} & 34.2 {\scriptsize\textcolor{gainred}{(-4.2\%)}} & 32.8 {\scriptsize\textcolor{gainred}{(-7.4\%)}} \\
& w/o DW & 33.1 {\scriptsize\textcolor{gainred}{(-5.5\%)}} & 32.5 {\scriptsize\textcolor{gainred}{(-6.2\%)}} & 31.7 {\scriptsize\textcolor{gainred}{(-11.1\%)}} & 31.0 {\scriptsize\textcolor{gainred}{(-12.4\%)}} \\
\midrule

\multirow{3}{*}{WN18RR}
& {\method} & \textbf{49.0} & \textbf{25.9} & \textbf{49.4} & \textbf{46.0} \\
\cmidrule(lr){2-6}
& w/o RA & 47.8 {\scriptsize\textcolor{gainred}{(-2.5\%)}} & 23.8 {\scriptsize\textcolor{gainred}{(-8.4\%)}} & 47.7 {\scriptsize\textcolor{gainred}{(-3.6\%)}} & 44.5 {\scriptsize\textcolor{gainred}{(-3.1\%)}} \\
& w/o DW & 47.8 {\scriptsize\textcolor{gainred}{(-2.5\%)}} & 23.7 {\scriptsize\textcolor{gainred}{(-8.4\%)}} & 47.4 {\scriptsize\textcolor{gainred}{(-4.2\%)}} & 44.2 {\scriptsize\textcolor{gainred}{(-4.1\%)}} \\
\midrule

\multirow{3}{*}{YAGO3-10}
& {\method} & \textbf{50.8} & \textbf{51.8} & \textbf{54.5} & \textbf{53.0} \\
\cmidrule(lr){2-6}
& w/o RA & 50.0 {\scriptsize\textcolor{gainred}{(-1.6\%)}} & 50.5 {\scriptsize\textcolor{gainred}{(-2.5\%)}} & 50.8 {\scriptsize\textcolor{gainred}{(-6.9\%)}} & 48.5 {\scriptsize\textcolor{gainred}{(-8.3\%)}} \\
& w/o DW & 45.0 {\scriptsize\textcolor{gainred}{(-11.4\%)}} & 42.8 {\scriptsize\textcolor{gainred}{(-17.4\%)}} & 42.8 {\scriptsize\textcolor{gainred}{(-21.4\%)}} & 38.2 {\scriptsize\textcolor{gainred}{(-27.9\%)}} \\
\bottomrule
\end{tabular}
\end{table}

\subsection{EQ3. Ablation Study}\label{sec:eval-eq3}
In this experiment, we verify the effectiveness of the two proposed components of {\method}: 
risk-aware negative budget allocation (RA) for (C1) and distribution-calibrated negative weighting (DW) for (C2).
We compare {\method} with the following ablated variants:
\begin{itemize}[leftmargin=8pt]
\item \textbf{w/o RA}: uses a fixed negative budget ($\gamma=0$), while applying our distribution-calibrated weighting.
\item \textbf{w/o DW}: uses our risk-aware negative budget allocation, but assigns uniform weights to all negatives ($\alpha=0$).
\end{itemize}

\noindent
\textbf{Effectiveness of individual strategies.}
Table~\ref{tab:ablation} shows that removing either component degrades KGE performance in all model--dataset combinations, demonstrating that both RA and DW contribute to the effectiveness of {\method}.
Without RA, MRR decreases by 1.6--8.4\% (4.5\% on average), validating the benefit of adapting the negative budget to false-negative risk.
Without DW, MRR decreases by 2.5\%--27.9\% (11.0\% on average), with the larger degradation indicating that 
effective weighting remains crucial even after the negative budget is adapted to reduce false-negative risk.

\begin{figure}[t]
\centering
{\footnotesize
\legendbar{legend1}\hspace{0.35em}w/o RA (vs. Self-Adv)
\quad\quad
\legendbar{legend2}\hspace{0.35em}w/o DW (vs. Uniform)
}
\vspace{0.4em}

\includegraphics[width=0.9\linewidth]{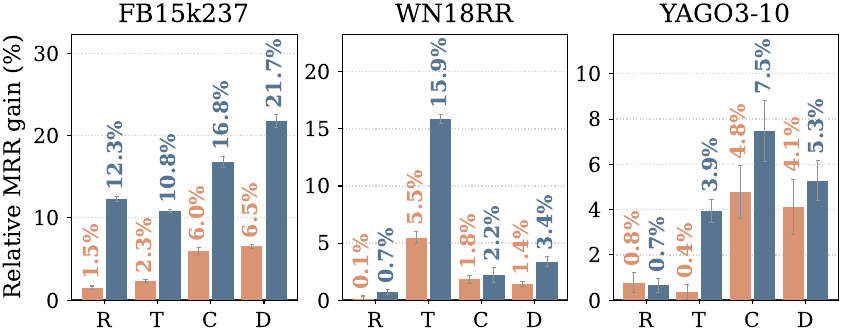}
\vspace{-1mm}
\caption{Relative MRR gains over the corresponding counterparts, {\sa} and Uniform. R, T, C, and D denote RotatE, TransE, ComplEx, and DistMult, respectively.}
\label{fig:ablation-generalized}
\vspace{-5mm}
\end{figure}

\vspace{0.5mm}
\noindent
\textbf{Relative performance over existing counterparts.}
We further analyze the effectiveness of each proposed strategy by comparing its relative performance against the corresponding existing counterpart.
Specifically, \textbf{w/o RA} uses the same fixed budget as Self-Adv but retains our DW, thereby isolating the effect of distribution-calibrated weighting.
Conversely, \textbf{w/o DW} uses the same uniform weighting as Uniform but retains our RA, thereby isolating the effect of risk-aware budget allocation.
As shown in Figure~\ref{fig:ablation-generalized}, \textbf{w/o RA} consistently outperforms Self-Adv, showing that incorporating score concentration in addition to dispersion provides more effective negative weighting.
This supports our motivation that dispersion and concentration capture complementary characteristics of negative-score distributions.
Moreover, \textbf{w/o DW} yields even larger gains over Uniform across all model--dataset combinations, despite using the same uniform weighting.
This demonstrates that adapting the negative budget according to false-negative risk alone provides substantial performance gains, further validating the importance of risk-aware negative budget allocation.

\subsection{EQ4. Sensitivity Analysis}\label{sec:eval-eq4}
In this experiment, we analyze the sensitivity of {\method} to its hyperparameters, $\gamma$, $\lambda$, and $\beta$.
For each analysis, we vary one hyperparameter while fixing the others to their default values.
Detailed settings and additional sensitivity analyses for the adjustment parameters $\tau$ and $\omega$ are provided in Appendix~\ref{app:sensitivity}.

\begin{figure}
\centering
\modellegend

\setlength\tabcolsep{0.2pt}
\def\arraystretch{0.5}
\begin{tabular}{ccc}
\includegraphics[width=0.31\linewidth]{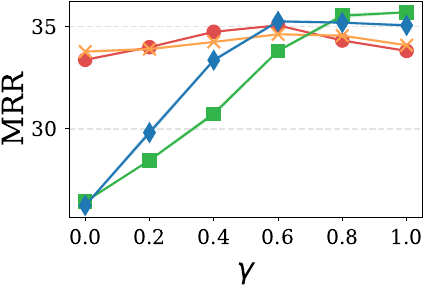} & 
\includegraphics[width=0.31\linewidth]{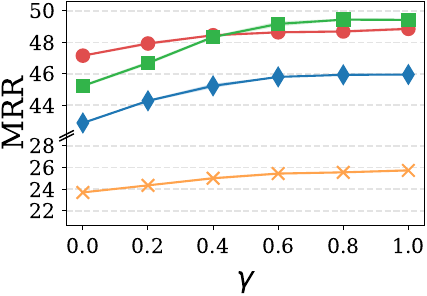} & 
\includegraphics[width=0.31\linewidth]{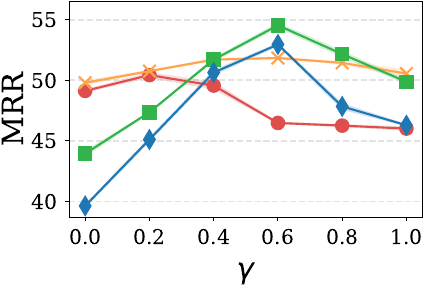} \\
\includegraphics[width=0.31\linewidth]{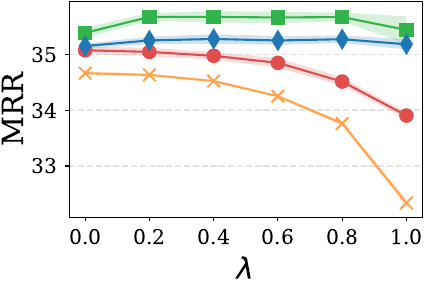} & 
\includegraphics[width=0.31\linewidth]{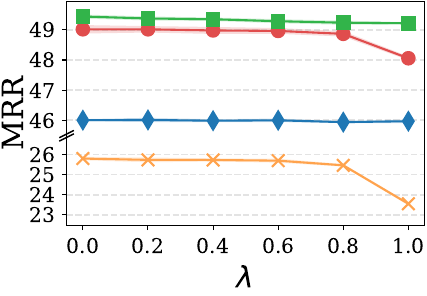} & 
\includegraphics[width=0.31\linewidth]{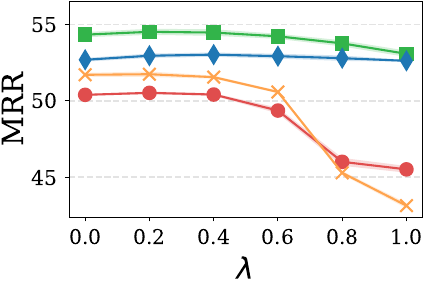} \\
\includegraphics[width=0.31\linewidth]{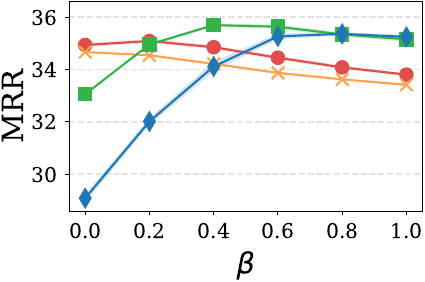} & 
\includegraphics[width=0.31\linewidth]{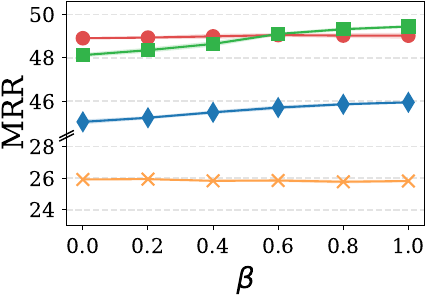} & 
\includegraphics[width=0.31\linewidth]{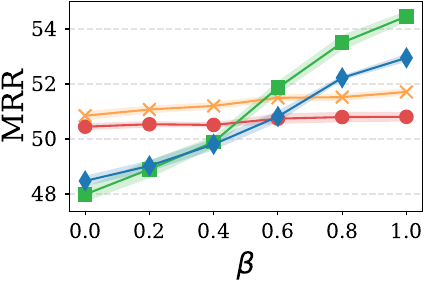} \\
(a) FB15k237 & (b) WN18RR & (c) YAGO3-10 \\
\end{tabular}
\vspace{-1mm}
\caption{Hyperparameter sensitivity analysis of {\method}.}
\label{fig:hyperparameter}
\vspace{-5.5mm}
\end{figure}

\noindent
\textbf{Sensitivity to $\gamma$.}
The hyperparameter $\gamma$ controls the degree of adapting the negative budget to false-negative risk, where $\gamma=0$ corresponds to a fixed negative budget and a larger $\gamma$ reduces the budget for higher-risk positives more.
As shown in the top row of Figure~\ref{fig:hyperparameter}, the best performance is consistently achieved with $\gamma>0$ across all KGE models and datasets, validating the benefit of adapting the negative budget according to false-negative risk rather than using a fixed budget.
Performance tends to improve as $\gamma$ increases, with the best results typically observed around $\gamma=0.6$--$1.0$.
On YAGO3-10, however, performance tends to peak around $\gamma=0.6$ and then decrease, suggesting that overly aggressive budget reduction can discard useful negative supervision.

\vspace{0.5mm}
\noindent
\textbf{Sensitivity to $\lambda$.}
The hyperparameter $\lambda$ balances score dispersion $\sigma_\x$ and concentration $C_\x$ in our distribution-calibrated weighting, where $\lambda=0$ considers only dispersion and $\lambda=1$ considers only concentration.
Across most settings, performance initially improves as $\lambda$ increases and then decreases at larger values, with the best performance generally achieved at intermediate values.
This indicates that neither dispersion nor concentration alone is sufficient and supports our design of jointly considering both characteristics when determining negative weights.

\vspace{0.5mm}
\noindent
\textbf{Sensitivity to $\beta$.}
The hyperparameter $\beta$ controls budget-aware weight normalization, where $\beta=0$ applies no normalization and increasing $\beta$ compensates for the larger average per-negative weights induced by smaller budgets.
Performance tends to improve as $\beta$ increases, indicating that assigning excessively large weights to negatives from smaller-budget, higher-risk positives can be detrimental.
This result supports the necessity of our normalization strategy for mitigating unintended weight amplification caused by adaptive negative budgets.


\section{Conclusion and Future Work}\label{sec:con}
In this work, we identify two fundamental yet underexplored challenges in contrastive learning with negative samples for KGE: \textbf{(C1)} \textit{negative budget allocation} and \textbf{(C2)} \textit{negative importance estimation}.
To effectively address these challenges, we propose {\method}, which employs (1) \textit{risk-aware negative budget allocation} and (2) \textit{distribution-calibrated negative weighting}.
The former adaptively determines the negative budget based on false-negative sampling risk, while the latter calibrates negative contributions to learning based on two complementary characteristics of the negative-score distribution: dispersion and concentration.
Extensive experiments with four KGE models on three real-world KGs demonstrate that {\method} (1) achieves the best average KGC performance, (2) improves training efficiency, and (3) benefits from both proposed strategies, which provide complementary gains.

While these results demonstrate the effectiveness of {\method}, several aspects remain open for further works.
Our analysis (Section~\ref{sec:proposed-owa}) assumes uniform observability and shared entity and relation sets between open- and closed-world KGs; extending it to heterogeneous observability and inductive settings with unseen entities or relations is an important direction for future work.
Moreover, while {\method} focuses on determining the negative budget and weights, jointly optimizing the negative sampling strategy itself could further improve negative selection.



\bibliographystyle{plainnat}
\balance
\bibliography{bibliography.bib}

\appendix
\section{Appendix}

\subsection{Implementation Details}\label{app:implementation-details}
The implementation details of our experiments are as follows:
\begin{itemize}[leftmargin=10pt]

\item \textbf{System configurations:}
We run all experiments on a machine with an NVIDIA RTX 5000 PRO GPU, with dual Intel Xeon Gold 6526Y CPUs (32 physical cores, 64 threads in total) with 256 GB of main memory.

\item \textbf{Training:}
We tune hyperparameters and select the configuration with the highest validation MRR. 
Each model is trained for 100K steps and evaluated every 5K steps. 
We decay the learning rate by 10\% after 50K steps. 
For a fair comparison, all baselines are trained under the same hardware environment and memory configuration.
For reproducibility, we release the code, datasets, and complete hyperparameter settings at {\codeurl}.

\item \textbf{Evaluation:}
We adopt the standard filtered setting~\citep{bordes2013translating} alongside the tie-breaking protocol from \citet{sun2020reevaluation} when computing rank-based metrics. 
Under the filtered setting, candidate triples that exist as ground-truth facts in the dataset (other than the target entity) are excluded during ranking. 
Under the tie-breaking protocol, if $n$ entities obtain a strictly higher score than the target entity and $m$ entities receive the same score, the final rank is calculated as $n + \frac{m+1}{2}$ in order to resolve identical scores fairly and prevent overestimation~\cite{sun2020reevaluation}.

\end{itemize}

\subsection{Description of Datasets}\label{app:datasets}
We use the following three real-world KGs in our experiments. Table~\ref{tab:dataset} summarizes their statistics.
\begin{itemize}[leftmargin=8pt]
\item \textbf{FB15k237}~\cite{toutanova2015observed}: A subset of Freebase~\cite{bollacker2008freebase} containing real-world facts across diverse domains, such as people, locations, and organizations. It removes inverse relations from FB15k to prevent information leakage.

\item \textbf{WN18RR}~\cite{dettmers2018convolutional}: A subset of WordNet, a lexical knowledge base representing semantic relationships between words, such as hypernymy and derivational relations. It removes inverse relations from WN18.

\item \textbf{YAGO3-10}~\cite{dettmers2018convolutional}: A subset of YAGO3~\cite{suchanek2007yago,mahdisoltani2013yago3}, a large-scale knowledge graph primarily constructed from Wikipedia, containing factual relationships among real-world entities.
\end{itemize}
\begin{table}[h]
\centering
\setlength{\tabcolsep}{5pt}
\def\arraystretch{0.8} 
\caption{Statistics of three KGs used in our experiments.}
\begin{tabular}{c|ccc}
\toprule
 & $|\E|$ & $|\R|$ & $|\T|$ \\
\midrule
FB15k237 & 14,541 & 237 & 272,115 \\
WN18RR & 40,943 & 11 & 86,835 \\
YAGO3-10 & 123,182 & 37 & 1,079,040 \\
\bottomrule
\end{tabular}
\label{tab:dataset}
\end{table}

\subsection{Description of Baselines}\label{app:baselines}
We compare 15 baselines with {\method} in our experiments.
\begin{itemize}[leftmargin=8pt]
\item \textbf{Uniform Sampling}~\cite{bordes2013translating} randomly corrupts the head or tail entity of a positive triple using a uniformly sampled entity. While efficient, it often produces trivial negatives.

\item \textbf{KBGAN}~\cite{cai2018kbgan} employs an adversarial framework where a generator selects challenging negatives for a discriminator.

\item \textbf{NSCaching}~\cite{zhang2019nscaching} maintains dynamically updated caches of high-scoring negatives, balancing exploration and exploitation.

\item \textbf{\sa}~\cite{sun2019rotate} weights uniformly sampled negatives using a temperature-controlled softmax over model scores, emphasizing harder negatives.


\item \textbf{SANS} and \textbf{RW-SANS}~\cite{ahrabian2020sans} sample structurally relevant negatives from $k$-hop neighborhoods and their random-walk approximations, respectively. We combine each with Uniform and {\sa} weighting, resulting in four more variants: Uniform SANS, Uniform RW-SANS, {\sa} SANS, and {\sa} RW-SANS.

\item \textbf{EANS}~\cite{je2022eans} exploits embedding similarity by reorganizing entity indices and applying Gaussian sampling to select nearby negatives, with an auxiliary loss for false-negative mitigation.

\item \textbf{ERDNS}~\cite{yao2023erdns} allocates a fixed negative budget via multinomial sampling based on entity--relation frequencies, motivated by optimal embedding analysis~\cite{kamigaito2022comprehensive}.

\item \textbf{DeMix-Adv}~\cite{chen2023demix} distinguishes suspected false negatives from reliable negatives, mixing the former with positives and the latter with other negatives to generate informative samples.

\item \textbf{M$^2$ixKG}~\cite{che2024m2ixkg} synthesizes hard negatives through embedding interpolation, mixing entities from relation-sharing triples or high-scoring negatives.

\item \textbf{IF-NS}~\cite{cai2025ifns} uses influence functions to identify informative negatives and maintains a dynamically updated candidate repository.

\item \textbf{BigramKG}~\cite{che2025bigramkg} combines relation--entity semantic matching probabilities with model scores to emphasize hard negatives and suppress potential false negatives.

\item \textbf{ConDNS}~\cite{wang2025condns} uses conditional diffusion to generate synthetic negatives in embedding space, controlling difficulty through denoising timesteps.
\end{itemize}

\subsection{Sensitivity Analysis: $\tau$ and $\omega$}\label{app:sensitivity}
As described in Section~\ref{sec:proposed-q1-budget}, we further consider two strategies for adjusting the estimated risk: 
(a) uncertainty-aware and (b) relation-aware adjustments.
(a) is controlled by $\tau$, where $\tau=0$ directly uses the observed anchor degree $d(\x)$ and a larger $\tau$ incorporates greater uncertainty into $d(\x)$.
(b) is controlled by $\omega$, where $\omega=0$ directly uses $d(\x)$ and a larger $\omega$ places greater emphasis on the relation-level degree.

\begin{figure}[t]
    \centering
    \modellegend\\
    \includegraphics[width=0.46\linewidth]{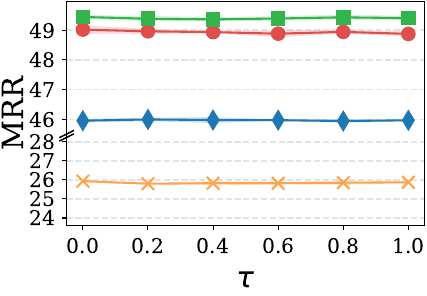}
    \includegraphics[width=0.46\linewidth]{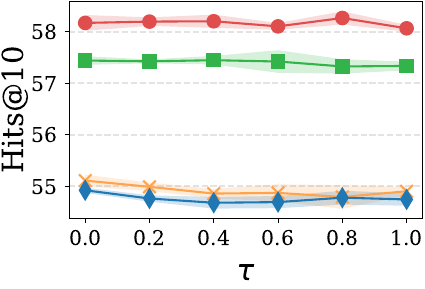} \\
    \includegraphics[width=0.46\linewidth]{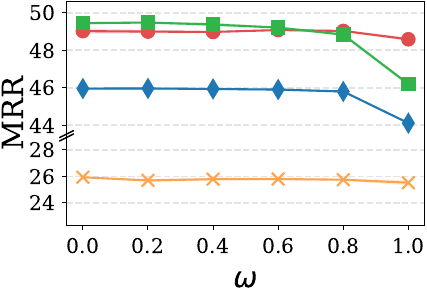}
    \includegraphics[width=0.46\linewidth]{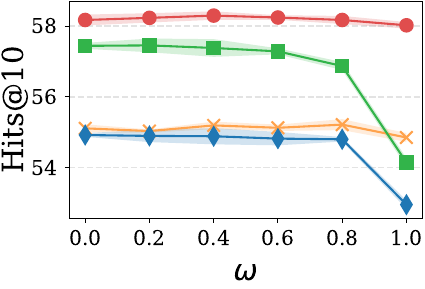} \\
    \caption{Sensitivity of KGE performance to the two adjustment parameters, $\tau$ and $\omega$, on WN18RR.}
    \label{fig:adjustments}
\end{figure}

Figure~\ref{fig:adjustments} shows the sensitivity of the KGE performance (MRR and H@10) to $\tau$ and $\omega$.
For $\tau$, performance remains largely stable across a wide range of values, indicating that {\method} is highly insensitive to the assumed uncertainty in $d(\x)$.
This suggests that the observed anchor degree, $d(\x)$, itself provides a sufficiently reliable estimate for risk-aware negative budget allocation.
For $\omega$, performance also remains stable up to around $\omega=0.8$, showing that incorporating relation-level information yields comparable performance to directly using $d(\x)$.
However, performance degrades when $\omega>0.8$, indicating that excessive reliance on relation-level information can obscure anchor-specific risk.
Overall, these results demonstrate that the original degree-based design of {\method} is sufficiently reliable and {\method} is insensitive to the adjustment hyperparameters.


\subsection{Relation Category Analysis}\label{app:relation-category}
We explain how relations are categorized into four types: 1-to-1, 1-to-N, N-to-1, and N-to-N. 
Following \citet{wang2014transh}, for each relation $r$, we compute the average number of tails per head (\textit{tphr}) and the average number of heads per tail (\textit{hptr}) from the training data.
We classify relations as follows: 
1-to-1 if \textit{tphr} $<1.5$ and \textit{hptr} $<1.5$; 
1-to-N if \textit{tphr} $\geq1.5$ and \textit{hptr} $<1.5$; 
N-to-1 if \textit{tphr} $<1.5$ and \textit{hptr} $\geq1.5$; 
and N-to-N otherwise.
Statistics of relation categories are summarized in Table~\ref{tab:rel-stat}.

\begin{table}[h]
\centering
\footnotesize
\def\arraystretch{0.9}
\setlength\tabcolsep{1pt}
\caption{Statistics of relation cardinality.}
\begin{tabular}{cccccc}
\toprule
Dataset & Statistic & 1-to-1 & 1-to-N & N-to-1 & N-to-N \\
\midrule
\multirow[c]{3}{*}{FB15k237}
 & \# Relations & 17 & 26 & 86 & 108 \\
 & \# Train triples & 4,278 (1.6\%) & 12,536 (4.6\%) & 50,635 (18.6\%) & 204,666 (75.2\%) \\
 & \# Test triples & 192 (0.9\%) & 1,293 (6.3\%) & 4,508 (22.0\%) & 14,473 (70.7\%) \\
\midrule
\multirow[c]{3}{*}{WN18RR}
 & \# Relations & 2 & 4 & 3 & 2 \\
 & \# Train triples & 1,218 (1.4\%) & 13,770 (15.9\%) & 40,833 (47.0\%) & 31,014 (35.7\%) \\
 & \# Test triples & 42 (1.3\%) & 475 (15.2\%) & 1,487 (47.4\%) & 1,130 (36.1\%) \\
\midrule
\multirow[c]{3}{*}{YAGO3-10}
 & \# Relations & 2 & 5 & 10 & 20 \\
 & \# Train triples & 6,323 (0.6\%) & 15,356 (1.4\%) & 127,635 (11.8\%) & 929,726 (86.2\%) \\
 & \# Test triples & 30 (0.6\%) & 89 (1.8\%) & 591 (11.8\%) & 4,290 (85.8\%) \\
\bottomrule
\end{tabular}
\label{tab:rel-stat}
\end{table}

\begin{figure*}[t]
    \centering
    \begin{tabular}{c}        
    \includegraphics[width=0.93\textwidth]{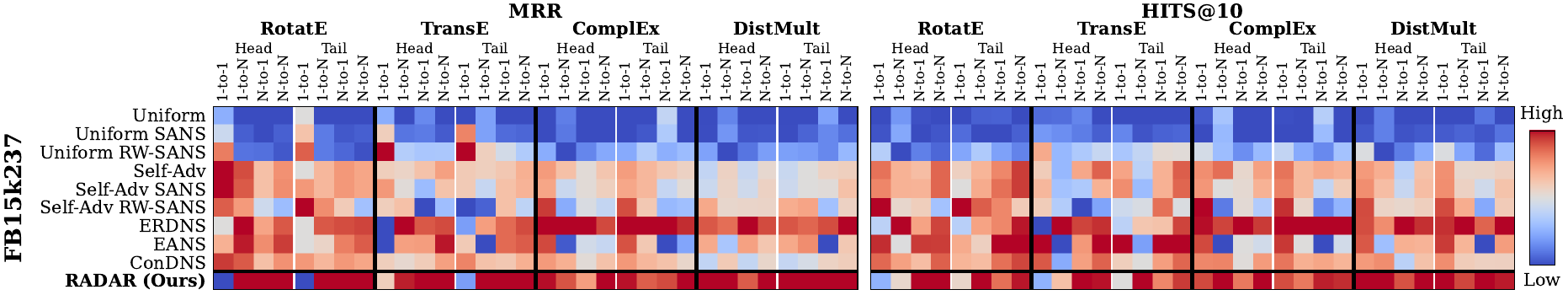} \\    
    \includegraphics[width=0.93\textwidth]{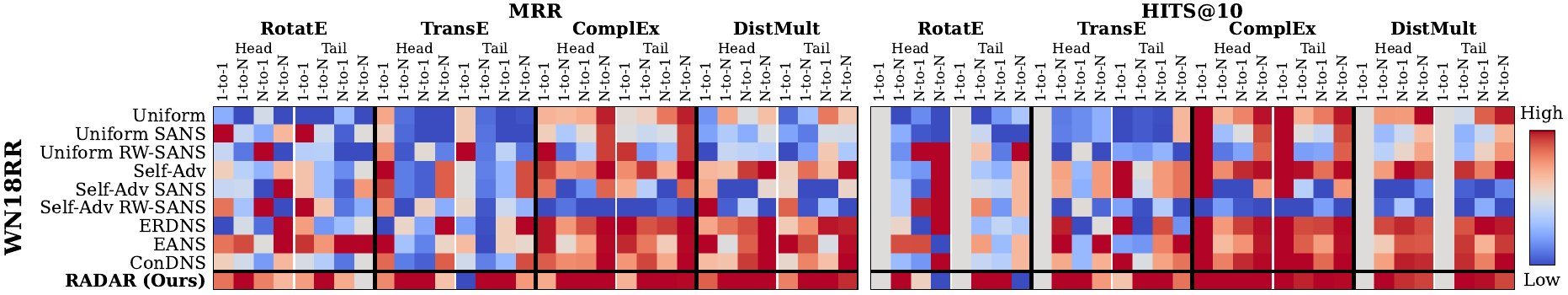} \\   
    \includegraphics[width=0.93\textwidth]{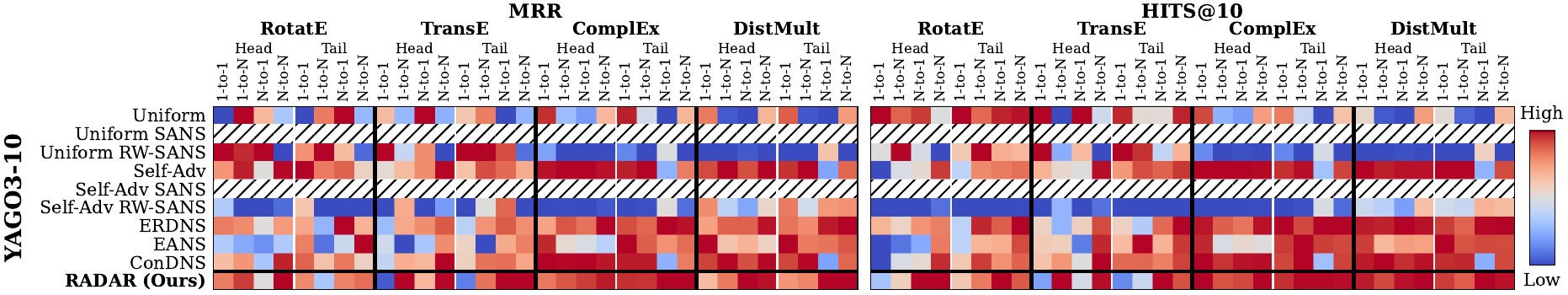} 
    \end{tabular}
    \caption{Negative sampling methods evaluated on different relation categories.}
    \label{fig:rel-category}
\end{figure*}

Figure~\ref{fig:rel-category} shows the complete results across all KGE model--dataset combinations, complementing the relation-cardinality analysis in EQ1 (Section~\ref{sec:eval-eq1}).
Overall, {\method} performs strongly across different relation categories, with particularly consistent gains on 1-to-N, N-to-1, and N-to-N relations.
These results further support the effectiveness of risk-aware negative budget allocation for multi-valued relations, where false-negative risk is higher.
In contrast, {\method} shows limited gains on 1-to-1 relations compared to baselines.
Nevertheless, as shown in Table~\ref{tab:rel-stat}, 1-to-1 relations account for only around $1\%$ of train/test triples across the three datasets and thus have limited impact on overall KGE performance.



\subsection{Performance--Efficiency Trade-Off}\label{app:training-efficiency}
\begin{figure*}[t]
    \centering
    \methodlegend
    
    \setlength{\tabcolsep}{1pt}
    \def\arraystretch{0.2} 
    \begin{tabular}{cccc}
    \includegraphics[width=0.21\linewidth]{figs/RotatE_FB15k237_t2p_MRR_log.pdf} &
    \includegraphics[width=0.21\linewidth]{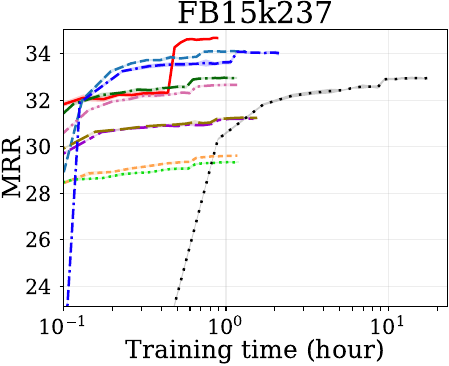} &
    \includegraphics[width=0.21\linewidth]{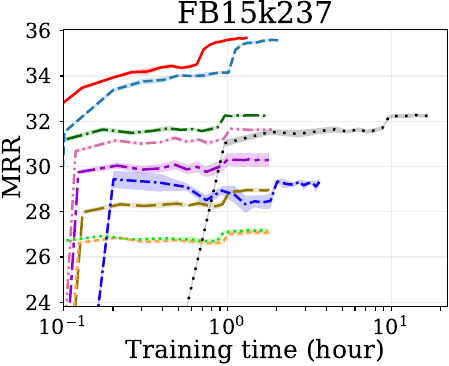} &
    \includegraphics[width=0.21\linewidth]{figs/DistMult_FB15k237_t2p_MRR_log.pdf} \\
    \includegraphics[width=0.21\linewidth]{figs/RotatE_wn18rr_t2p_MRR_log.pdf} &
    \includegraphics[width=0.21\linewidth]{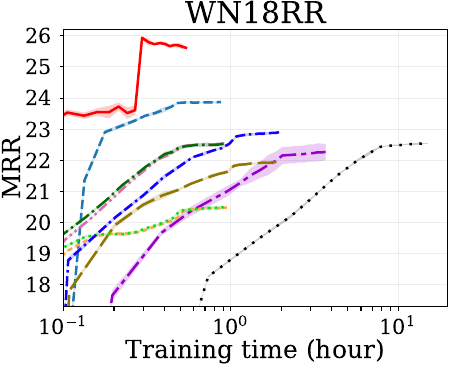} &
    \includegraphics[width=0.21\linewidth]{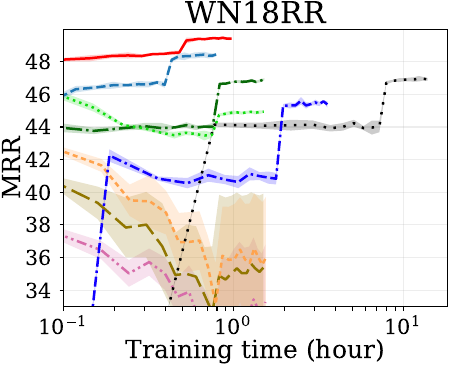} &
    \includegraphics[width=0.21\linewidth]{figs/DistMult_wn18rr_t2p_MRR_log.pdf} \\
    \includegraphics[width=0.21\linewidth]{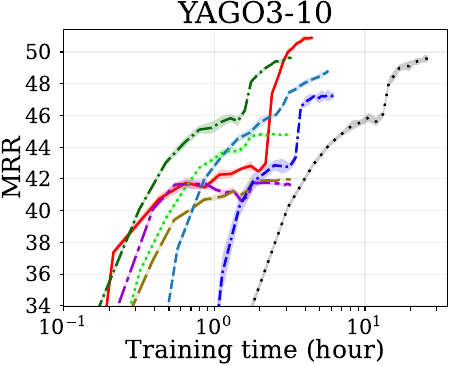} &
    \includegraphics[width=0.21\linewidth]{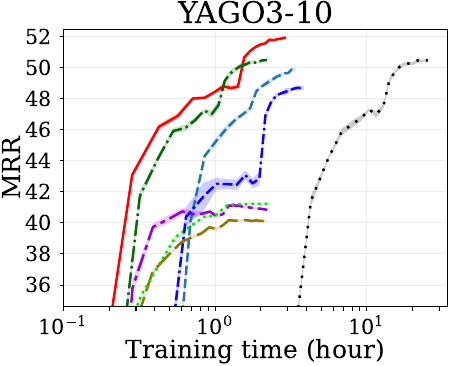} &
    \includegraphics[width=0.21\linewidth]{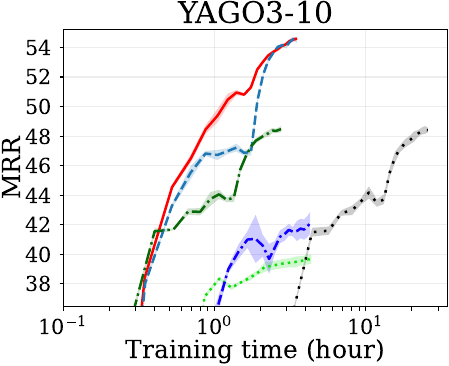} &
    \includegraphics[width=0.21\linewidth]{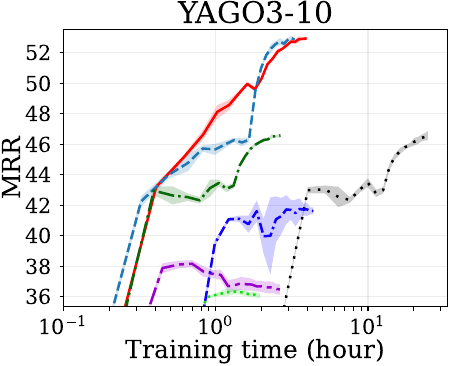} \\
    (a) RotatE & (b) TransE & (c) ComplEx & (d) DistMult 
    \end{tabular}
    \caption{Performance--efficiency trade-off of {\method} and baselines across all KGE models and datasets.}
    \label{fig:rel-cardinality-all}
\end{figure*}

Figure~\ref{fig:rel-cardinality-all} presents the complete performance--efficiency results across all KGE models and datasets, complementing the results in EQ2 (Section~\ref{sec:eval-eq2}).
Overall, {\method} exhibits a favorable performance--efficiency trade-off across diverse settings.
On FB15k237 and WN18RR, {\method} achieves the highest KGE performance in the shortest training time than other methods.
In particular, although {\method} is not always the fastest on YAGO3-10, it achieves the highest performance, demonstrating that its efficiency gains do not come at the cost of KGE performance.
These results further support the effectiveness of {\method} across different models and datasets.



\end{document}